\documentclass[11pt]{article}

\usepackage[preprint]{acl}
\usepackage{times}
\usepackage{latexsym}

\usepackage[T1]{fontenc}

\usepackage[utf8]{inputenc}

\usepackage{microtype}

\usepackage{inconsolata}

\usepackage{graphicx}

\usepackage{enumitem}
\usepackage{algorithm}
\usepackage{algpseudocode}
\usepackage{amsmath}
\usepackage{amsfonts,amssymb}
\usepackage{xspace}

\newcommand{\method}{SAT\xspace}

\algnewcommand{\Initialize}[1]{%
  \State \textbf{Initialize:}
  \Statex \hspace*{\algorithmicindent}\parbox[t]{.8\linewidth}{\raggedright #1}
}

\usepackage{booktabs}
\usepackage{multirow}
\usepackage{xcolor}
\usepackage{colortbl}
\usepackage{graphicx}
\usepackage{array}
\usepackage{tcolorbox}

\definecolor{myyellow}{RGB}{184, 80, 40}

\definecolor{myblue}{RGB}{46, 117, 145}

\title{\method: Balancing Reasoning Accuracy and Efficiency with Stepwise Adaptive Thinking}

\author{
  Weiyang Huang\textsuperscript{1},
  Xuefeng Bai\textsuperscript{1}\thanks{Corresponding author},
  Kehai Chen\textsuperscript{1}, Xinyang Chen\textsuperscript{1}, \\
  \textbf{Yibin Chen}\textsuperscript{2},
  \textbf{Weili Guan}\textsuperscript{1},
  \textbf{Min Zhang}\textsuperscript{1} \\
  \textsuperscript{1} Harbin Institute of Technology (Shenzhen), Shenzhen, China \\
  \textsuperscript{2} Huawei Technologies \\
}

\begin{document}
\maketitle
\begin{abstract}
Large Reasoning Models (LRMs) have revolutionized complex problem-solving, yet they exhibit a pervasive ``overthinking'', generating unnecessarily long reasoning chains.
While current solutions improve token efficiency, they often sacrifice fine-grained control or risk disrupting the logical integrity of the reasoning process. 
To address this, we introduce \textbf{S}tepwise \textbf{A}daptive \textbf{T}hinking (\textbf{SAT}), a framework that performs step-level, difficulty-aware pruning while preserving the core reasoning structure. 
SAT formulates reasoning as a Finite-State Machine (FSM) with distinct thinking modes (\textsc{Slow, Normal, Fast, Skip}). It navigates these states dynamically using a lightweight Process Reward Model (PRM), compressing easy steps while preserving depth for hard ones.
Experiments across 9 LRMs and 7 benchmarks show that SAT achieves up to 40\% reduction in reasoning tokens while generally maintaining or improving accuracy\footnote{Code is available on \href{https://github.com/byxw13/SAT_Code}{GitHub}.}
\end{abstract}

\section{Introduction}

The advent of large reasoning models (LRMs) such as OpenAI-O1~\citep{openai2025learning} and DeepSeek-R1~\citep{ds_qwen} has marked a significant advance in tackling complex reasoning tasks~\citep{wang-etal-2025-self-reasoning}. Nevertheless, these models are prone to a notable ``overthinking'' issue: they tend to produce excessively long Chains of Thought (COT) even for straightforward questions, creating a practical bottleneck in real-time or resource-constrained applications~\citep{sui2025stop}.

\begin{figure}[t]
    \centering
    \includegraphics[width=1.0\linewidth]{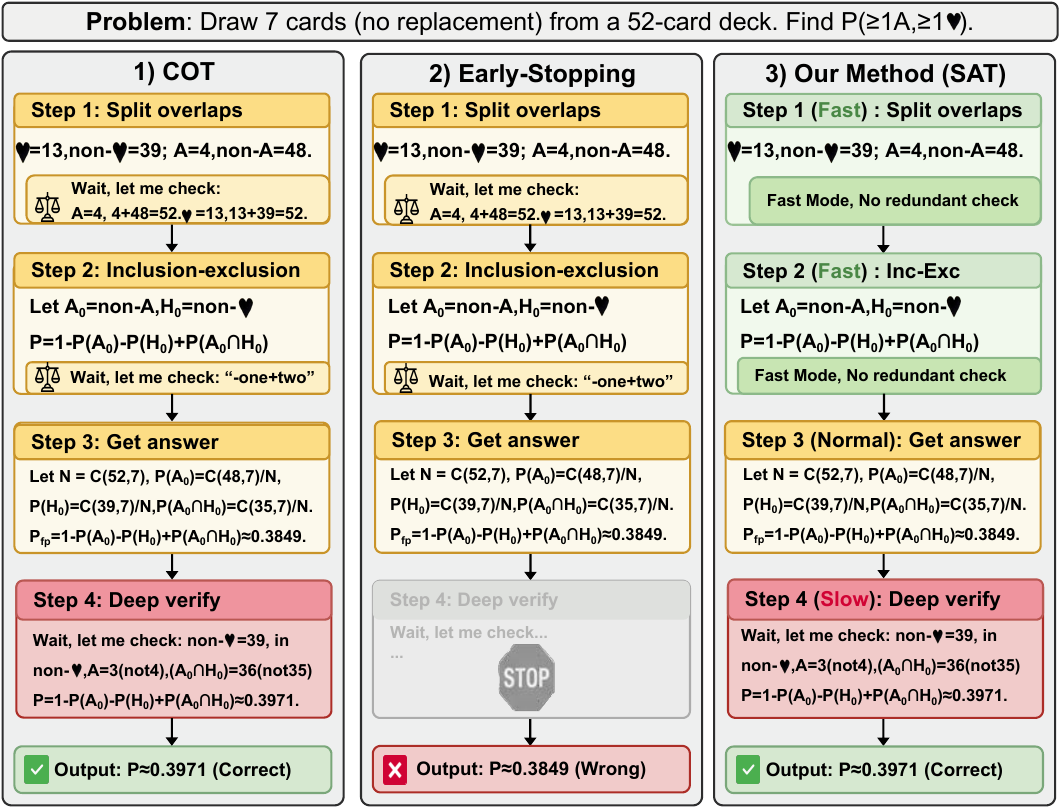}
    \caption{\textbf{Baselines vs. SAT.}
    COT spends redundant checks for easy steps, Early-Stopping halts after a high-confidence first-pass answer and fails, while \method skips redundancy on easy steps but preserves verification on hard steps, achieving the correct answer efficiently.}
    \label{fig:motivation}
\end{figure}

To mitigate this problem, numerous studies have sought to steer LLMs toward more token-efficient reasoning.
Earlier work employs strategic prompting~\cite{ccot,chen2024unlocking,xu2025chaindraftthinkingfaster}, question-driven routing~\cite{thinkswitcher,DiffAdapt}, planning and budgeting~\cite{tale,lin2025planbudgeteffectiveefficient}. 
However, these methods often regulate the reasoning process at a coarse-grained level, offering limited precise control over both the reasoning flow and its efficiency.
A more recent and promising direction explores dynamic decoding mechanisms, such as early stopping of the COT sequence~\cite{answer-early-stop,deer}, or suppressing reflective steps~\cite{huang2026efficient,wait}.
While effective in shortening COT, these methods rely on a uniform policy across all steps, ignoring their varying difficulty and necessity. As shown in Figure~\ref{fig:motivation}, this strategy risks disrupting or cutting off essential steps, thereby compromising logical integrity and overall performance.

\begin{figure*}[t] 
    \centering
    \includegraphics[width=0.8\linewidth]{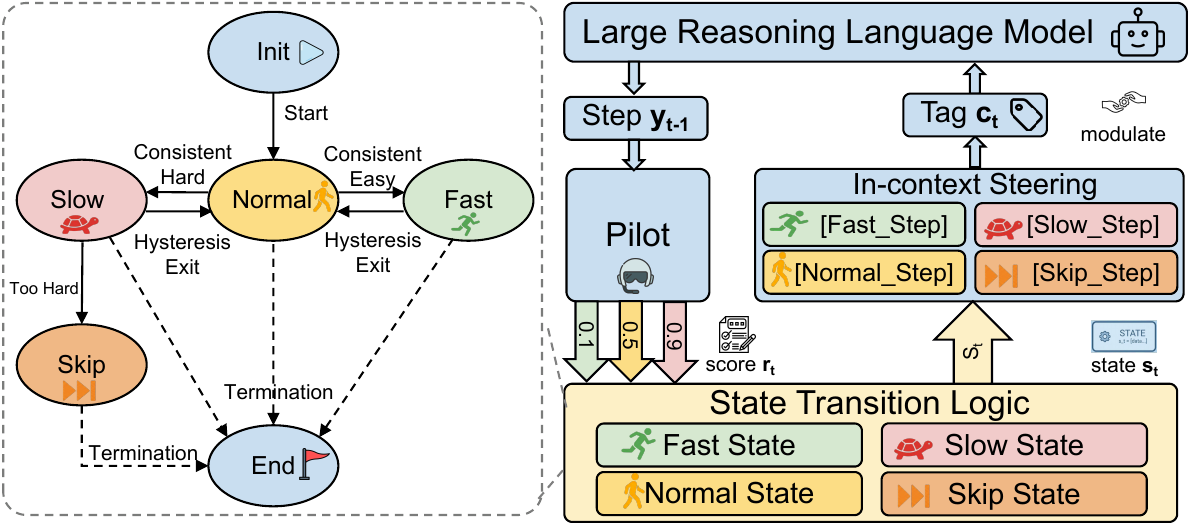} 
    
    \caption{\textbf{Overview of the proposed framework that models LRM reasoning as a Finite-State Machine (FSM).} 
    The left panel provides a detailed view of the specific state transition dynamics and stability rules of the proposed reasoning FSM.
    The right panel illustrates the closed-loop control flow: the \textbf{Pilot} perceives the difficulty ($r_t$) of the previous step ($y_{t-1}$), driving the \textbf{State Transition Logic} to update the thinking state ($s_{t}$). 
    This state is then mapped to a control tag ($c_{t}$) via \textbf{In-Context Steering} to modulate the next generation step ($y_{t}$). 
    }
    \label{fig:framework}
\end{figure*}

In this work, we propose~\method~(Stepwise Adaptive Thinking), a step-level, difficulty-aware pruning that eliminates redundancy while rigorously preserving the core logical structure essential for correct answers. 
As shown in Figure~\ref{fig:motivation}, departing from one-size-fits-all strategies, SAT is premised on the observation that reasoning steps within a solution vary in difficulty~\cite{saha2025system}.\footnote{See Appendix~\ref{app:intro_motivation} for comparisons with suppressing reflective methods.} Consequently, it dynamically navigates LLMs to employ a deeper thinking mode for challenging steps and a shallower, more efficient mode for simpler ones.
Specifically, as illustrated in Figure~\ref{fig:framework}, \method formulates the reasoning process as a Finite-State Machine (FSM) wherein each reasoning step is intrinsically linked to an intermediate thinking state categorized into one of four distinct thinking modes: \textsc{Slow}, \textsc{Normal}, \textsc{Fast}, and \textsc{Skip}. These modes represent varying degrees of reasoning depth and computational effort.
During inference, \method dynamically navigates through this state space based on the evolving historical reasoning context, allowing for the selective compression or deliberate elaboration of the COT sequence.

To estimate the appropriate thinking state at each step, we introduce a step-level difficulty score based on progress reward and then develop a lightweight Process Reward Model (PRM, \citealp{prm_origin}) for estimating the difficulty score based on confidence pattern and semantic information from the historical reasoning context. 
Leveraging these difficulty scores, \method dynamically steers the reasoning trajectory by transitioning between states: favoring concise modes such as \textsc{Fast} when the reasoning step is deemed ``easy'', and invoking more exhaustive modes like \textsc{Normal} and \textsc{SLOW} for ``complex'' steps, while resorting to \textsc{Skip} in overly difficult cases to guide the LLM toward an early, succinct conclusion. This control is achieved by injecting targeted prompting signals into the LLM input, directing it toward the desired reasoning depth.
The framework offers two key advantages: (1) it gains efficiency from pruning redundancy without compromising the reasoning depth essential for correctness; (2) it functions as a lightweight, real-time ``navigator'' for the LLM’s reasoning process, requiring no modifications to the base LLM's parameters and introducing minimal computational overhead.

We systematically evaluate the proposed method across 9 large reasoning models and 7 widely-used established benchmarks, covering mathematical reasoning, scientific reasoning, and programming tasks.
Experimental results show that \method achieves better efficiency-performance balance compared to strong baselines.
Further analysis demonstrates that \method incurs negligible overhead to achieve a 37\% end-to-end speedup, while exhibiting {difficulty-aware adaptivity} that dynamically aligns reasoning depth with step difficulty, generalizing effectively across diverse domains.

Our contributions are summarized as follows:
\begin{itemize}[itemsep=2pt,topsep=0pt,parsep=0pt,leftmargin=11pt]
    \item We propose \method, a {state machine-based} framework that prunes redundancy at the step-level without compromising reasoning completeness.
    \item We design a {lightweight} PRM that reduces parameters by 99\%, rendering real-time reasoning navigation practically feasible.
    \item Experiments on seven benchmarks demonstrate that our method achieves up to 40\% token compression while generally maintaining accuracy.
\end{itemize}

\section{Related Work}

\paragraph{Mitigating Overthinking.}
Existing strategies for mitigating the ``overthinking'' problem at inference time can be categorized into two groups, based on when and how they intervene in the reasoning process.
\textbf{Static Strategies} impose constraints or select paths before generation. CCOT~\citep{ccot} explicitly instructs the model to be concise using few-shot demonstrations;\citealt{chen2024unlocking} feed the model with pre-computed solution templates matched to the estimated task difficulty. 
{Question-driven routing}~\citep{thinkswitcher,zhang2024question} classify queries into discrete difficulty levels (e.g., simple, hard) at the onset to assign reasoning strategies.
{Planning and Budgeting}~\citep{tale} estimates a token budget beforehand and prompts the model to solve the problem within this limit.
However, these approaches rely on static, coarse-grained estimates that offer limited precise control over both the reasoning flow and its efficiency.
\textbf{Dynamic Interventions} intervene directly during inference~\citep{li2025generator}. 
{Early stopping methods}~\citep{answer-early-stop,deer,TrimR} seek to halt generation upon identifying moments of ``sufficient certainty'', while {suppression-based methods}~\citep{wait, huang2026efficient} mitigate redundancy by lowering the sampling probability of reflection-triggering tokens.
While adaptive, these methods tend to \textit{disrupt} the intrinsic reasoning process (e.g., truncating essential self-corrections or logical chains), degrading reasoning performance.

Instead of applying uniform compression or global early stopping, this work dynamically assesses the difficulty of each reasoning step and selectively prunes only redundant segments, thereby preserving the essential logical progression of the COT while achieving significant efficiency gains.

\paragraph{Process Reward Models.}

Process Reward Models (PRMs \citealt{prm_origin,qiyuan2025efficient}) have emerged as a pivotal technique for enhancing the reliability of reasoning in large models. By assigning quality scores to intermediate reasoning steps, they facilitate process‑supervised training of LRMs~\cite{luo2025ursa}. Beyond training, PRMs have also been effectively combined with inference‑time search strategies—such as Tree of Thoughts~\citep{prm_inference}, Best‑of‑N~\cite{DBLP:conf/nips/SunHZYQYWBZ24}, and beam search~\cite{PATS}—to prune erroneous reasoning paths and identify promising trajectories. 
However, PRM deployment is constrained by high computational cost~\citep{prm_limit}, as existing methods typically rely on external, heavyweight verifiers (e.g., 7B parameters). This makes them prohibitively expensive for efficient test‑time scaling.

In contrast, we leverage confidence patterns as the primary discriminative feature to construct a lightweight PRM (30M parameters) that reduces parameters by 99\% versus standard verifiers. This compact model is seamlessly integrated into our reasoning framework, where it monitors a single trajectory in real time and adaptively modulates reasoning density with negligible latency.

\section{Methodology}
\label{sec:fsm}

\paragraph{Problem Formulation.}
Given an input query $q$, a LRM produces a COT trajectory with $T$ steps, $Y=\{y_1,y_2,\dots,y_T\}$, and finally outputs an answer $a$.
The $t$-th step is generated solely conditioned on the preceding context:
\vspace{-0.5em}
\begin{equation}
    y_t \sim P(\cdot \mid q, y_{<t}).
    \label{eq:std-reasoning}
    \vspace{-0.5em}
\end{equation}
Following \citet{prm_origin}, each reasoning step $y_t$ is defined as a text segment delimited by a newline character \texttt{\textbackslash n} in the generated COT.
This static paradigm allocates a \textit{uniform thinking mode} to all steps, regardless of their difficulty, which often leads to ``overthinking'' for simpler steps. 

\subsection{Overview of the Reasoning FSM}
In this work, we frame the reasoning process as a Finite-State Machine (FSM) (Figure~\ref{fig:framework}). Unlike static approaches, this machine dynamically consumes the reasoning history and produces the subsequent reasoning steps. The reasoning FSM is defined as a hexad $\mathcal{M} = \langle \mathcal{S}, \Sigma, \Gamma, s_0, \delta, \omega \rangle$:

\begin{itemize}[itemsep=1pt,topsep=0pt,leftmargin=12pt]
    \item $\mathcal{S}$ is the set of \textbf{thinking states} (including the initial state $s_0{=}\texttt{INIT}$) regulating reasoning depth.
    \item $\Sigma$ is the \textbf{input space} of text segments; the input is the previous step $y_{t-1} \in \Sigma$.
    \item $\Gamma$ is the \textbf{output space} ($\Gamma \equiv \Sigma$), where the output is the current step $y_t$.
    \item $\delta: \mathcal{S} \times \Sigma \rightarrow \mathcal{S}$ is the \textbf{transition function}, driven by the Pilot to update the state based on the semantic difficulty of $y_{t-1}$.
    \item $\omega: \mathcal{S} \rightarrow \Gamma$ is the \textbf{emission function}, where the LRM generates the next step $y_t$ conditioned on the current state $s_t$.
\end{itemize}

\paragraph{State Space $\mathcal{S}$.}
We define the state space as $\mathcal{S}=\{\texttt{INIT}, \texttt{NORMAL}, \texttt{FAST}, \texttt{SLOW}, \texttt{SKIP}, \texttt{END}\}$.
Beyond the boundary states \texttt{INIT} and \texttt{END}, we design four modes to modulate the \textbf{emission function $\omega$}:
\texttt{NORMAL} maintains standard depth;
\texttt{FAST} accelerates simple steps by omitting redundancy;
\texttt{SLOW} handles complex steps via detailed thinking;
and \texttt{SKIP} acts as a ``soft'' termination, guiding the model to conclude naturally rather than rigid truncation.

\paragraph{State Transition $\delta$.}
At each step $t$, the transition function updates the machine's state by analyzing the step difficulty of the incoming text segment $y_{t-1}$.
Functionally, we decompose $\delta$ into a perception phase and a decision phase.
First, the Pilot serves as the perception kernel, mapping the textual input $y_{t-1}$ to a latent difficulty score $r_t \in [0,1]$.
Second, the state is updated based on this score and a history window $\mathcal{H}_k = \{r_{t-k+1}, \dots, r_{t-1}\}$:
\vspace{-0.5em}
\begin{equation}
    r_t = 1- \text{Pilot}(y_{t-1}),~s_t = f(s_{t-1}, r_t, \mathcal{H}_k).
\end{equation}
Section~\ref{subsec:difficulty} details the Pilot architecture, and Algorithm~\ref{alg:state_machine} outlines the specific update rules.

\paragraph{Emission Function $\omega$.}
This function governs the generation of the reasoning step $y_t$ by steering the LRM with a state-specific control token $c_t$:
\vspace{-0.5em}
\begin{equation}
    c_t = \text{Tag}(s_t), \quad y_t \sim P(\cdot \mid q, y_{<t}, c_t).
\end{equation}
The specific implementation of the tag mapping and steering prompts is detailed in Section~\ref{subsec:state-transition}.

\subsection{Pilot: Lightweight Stepwise Difficulty Estimation}
\label{subsec:difficulty}

\paragraph{Stepwise Difficulty Estimation.}
To instantiate the perception phase of the transition function $\delta$, we derive a difficulty score $r_t \in [0,1]$ for the historical steps.
Drawing on prior research that assesses problem-level difficulty based on the pass rate observed across multiple sampling trials~\citep{tong2024dartmath}, we extend this principle to the step level.
Specifically, we assess the difficulty by the likelihood that the current step leads to a correct final solution.
This objective aligns perfectly with the core function of Process Reward Models (PRMs)~\citep{Math-Shepherd}, which are explicitly designed to estimate this probability of correctness, denoted as $v_t$.
Consequently, we define the difficulty of the current step as the complement of its success probability:
\vspace{-0.5em}
\begin{equation}
    r_t = 1 - v_t.
\vspace{-0.5em}
\end{equation}
Intuitively, a lower probability of success (a lower PRM score) indicates a higher difficulty level, thereby necessitating deeper reasoning resources.

\paragraph{Lightweight Pilot.} Most existing PRMs are built on billion-parameter generative backbones (e.g., Qwen-2.5-7B), making them hard to be adapted to efficient reasoning. To address this, we introduce a 30M-parameter Pilot that removes generative overhead while retaining accurate difficulty estimation, yielding $\sim$99\% lower parameters than typical generative PRMs and retaining over 80\% of the performance of standard LLM-based PRMs.

\begin{figure}[t]
    \centering
    \includegraphics[width=0.6\linewidth]{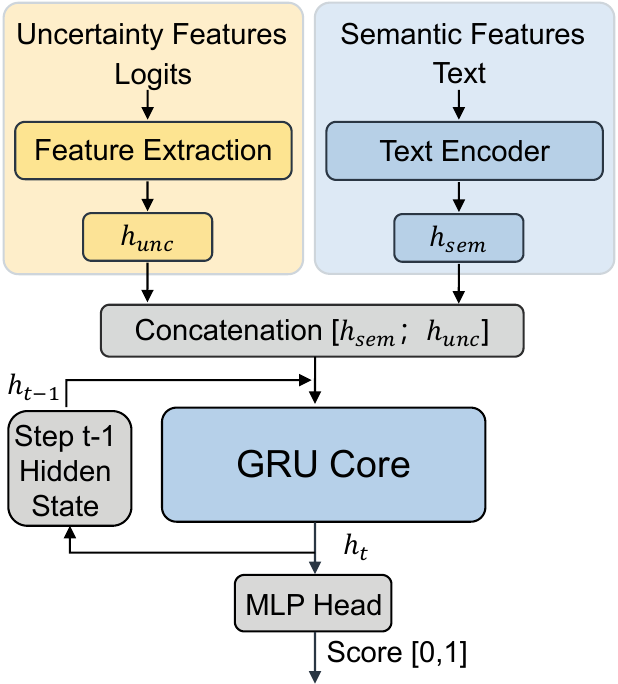} 
    \vspace{-0.3cm} 
    \caption{\textbf{Lightweight Pilot:} A GRU-based framework fusing semantic embeddings and uncertainty features to capture stepwise difficulty trajectories.}
    \label{fig:pilot_arch}
    \vspace{-0.5cm} 
\end{figure}

\paragraph{Architecture.}
As illustrated in Figure~\ref{fig:pilot_arch}, the Pilot maps the input $y_{t-1}$ into a fused representation $\mathbf{z}_t$ via two channels:
\textbf{Uncertainty Features ($\mathbf{h}_{\text{unc}}$)} derived from generation logits (e.g., entropy; see Appendix~\ref{app:input_features}) to quantify internal confidence, and
\textbf{Semantic Features ($\mathbf{h}_{\text{sem}}$)} obtained by encoding the text content via an encoder ({GTE-small}, \citealp{gte_small}).
These features are concatenated $\mathbf{z}_t=[\mathbf{h}_{\text{unc}};\mathbf{h}_{\text{sem}}]$ and processed by a GRU, projecting the hidden state to a scalar correctness score $v_t$ via learnable parameters $\mathbf{w}$ and $b$:
\vspace{-0.5em}
\begin{equation}
    \mathbf{h}_t = \text{GRU}(\mathbf{z}_t, \mathbf{h}_{t-1}), \quad v_t = \sigma(\mathbf{w}^\top \mathbf{h}_t + b).
    \vspace{-0.5em}
\end{equation}
The $v_t$ is converted to the difficulty score $r_t$, completing the perception phase of the state transition.

\paragraph{Teacher-Student Distillation.} We train the Pilot by distilling a teacher PRM. We generate step trajectories on {PRM800K}~\citep{prm_origin} using {DeepSeek-R1-Distill-Qwen-7B}~\citep{ds_qwen} and obtain teacher probabilities $v_t^{*}$ from {Skywork-o1-Open-PRM-7B}~\citep{Skywork-o1-Open-PRM-Qwen}. 
The Pilot is optimized with Binary Cross-Entropy to align its predicted correctness $v_t$:
\vspace{-0.5em}
\begin{equation} 
    \mathcal{L} = - \mathbb{E}\left[v_t^{*} \log v_t + (1-v_t^{*})\log(1-v_t)\right].
    \vspace{-0.5em}
\end{equation}
The distilled Pilot achieves high correlation with the teacher (as shown in Section~\ref{sec:ablation}) at orders-of-magnitude lower cost, enabling real-time feedback for online FSM control.
Crucially, since we freeze the gte encoder ($\sim$30M parameters) and only update the GRU ($\sim$0.15M parameters), the entire training process \textbf{converges in less than 5 minutes on a single GPU}.
This renders the deployment of \method highly accessible and orders of magnitude more efficient than training generative verifiers.

\subsection{Stepwise Difficulty Guided FSM}
\label{subsec:state-transition}

\begin{algorithm}[t]
    \small
    \caption{FSM Inference Step}
    \label{alg:state_machine}
    \begin{algorithmic}[1]
        
        \Statex \hspace{-\algorithmicindent}\parbox[t]{\dimexpr\linewidth+\algorithmicindent}{\textbf{Function:} Taking previous step $y_{t-1}$, state $s_{t-1}$, and history $\mathcal{H}$ as inputs, this algorithm executes one reasoning cycle: it perceives the difficulty of $y_{t-1}$, updates state $s_t$, and \textbf{generates the next reasoning step $y_t$} via the function $\omega$.}

        \State \textbf{if} $y_{t-1}$ contains \texttt{</think>} \textbf{then} \Return $(\texttt{END}, \varnothing)$
        
        \Statex \textbf{// 1. Perception Phase (Pilot)}
        \State $r_t \gets 1 - \text{Pilot}(y_{t-1})$ \Comment{Extract latent difficulty from text}
        \State Update history window $\mathcal{H}$ with $r_t$
        
        \Statex \textbf{// 2. Transition Logic (Dynamics)}
        \State $s_t \gets s_{t-1}$
        \State \textbf{if} $s_{t-1}=\texttt{INIT}$ \textbf{then} $s_t \gets \texttt{NORMAL}$

        \State \textbf{if} $s_{t-1} = \texttt{NORMAL}$ \textbf{then} \Comment{Consistent Entry}
        \State \hspace{\algorithmicindent} \textbf{if} $\text{All}(\mathcal{H} < \tau_{\texttt{fast}})$ \textbf{then} $s_t \gets \texttt{FAST}$
        \State \hspace{\algorithmicindent} \textbf{if} $\text{All}(\mathcal{H} > \tau_{\texttt{slow}})$ \textbf{then} $s_t \gets \texttt{SLOW}$

        \State \textbf{if} $s_{t-1} \in \{\texttt{FAST}, \texttt{SLOW}\}$ \textbf{then} \Comment{Hysteresis Exit}
        \State \hspace{\algorithmicindent} \textbf{if} Score crosses $\tau \pm \Delta$ \textbf{then} $s_t \gets \texttt{NORMAL}$

        \State \textbf{if} $s_t=\texttt{SLOW} \land \text{All}(\mathcal{H} > \tau_{\texttt{skip}})$ \textbf{then} $s_t \gets \texttt{SKIP}$ 
        
        \Statex \textbf{// 3. Emission Phase (Generation)}
        \State $c_t \gets \text{Tag}(s_t)$ \Comment{Determine control token}
        \State $y_t \gets \text{Generate}(q, y_{<t}, c_t)$ \Comment{Generate next step via $\omega$}
        \State \Return $(s_t, y_t)$
    \end{algorithmic}
    \vspace{-0.5em}
\end{algorithm}

Building on the stepwise difficulty score, we instantiate the FSM's \textbf{transition function $\delta$} with a difficulty-aware algorithm, and realize the \textbf{emission function $\omega$} via in-context steering.

\paragraph{Difficulty-aware Transition.} 
The transition logic is specified in Algorithm~\ref{alg:state_machine}. 
At each step boundary, the FSM first checks the termination condition. 
If the process continues, the state $s_t$ is updated based on the difficulty history $\mathcal{H}_k$ and a set of pre-defined \textbf{thresholds} $\mathcal{T} = \{\tau_{\texttt{fast}}, \tau_{\texttt{slow}}, \tau_{\texttt{skip}}\}$.
The logic follows two key principles:
(i) \textit{Consistent Entry}: The system transitions from \texttt{NORMAL} to \texttt{FAST} or \texttt{SLOW} only when the difficulty scores consistently breach the corresponding \textbf{threshold} (i.e., $\text{All}(\mathcal{H}_k < \tau_{\texttt{fast}})$).
(ii) \textit{Hysteresis Exit}: To prevent state flickering, returning to \texttt{NORMAL} requires the score $r_t$ to cross the \textbf{threshold} with a safety margin $\Delta$.
Additionally, if the difficulty remains excessively high while in the \texttt{SLOW} state, the state shifts to \texttt{SKIP} to encourage consolidation.

\paragraph{Emission via In-Context Steering.}
To implement the emission function $\omega$, we steer the generation by manipulating the input context of the LRM.
Specifically, we append the state-specific control tag $c_t = \text{Tag}(s_t)$ to the current reasoning context:
\vspace{-0.5em}
\begin{equation}
    \text{Context}_{t+1} = \text{Context}_{t} \oplus \text{Token}(c_t).
    \vspace{-0.5em}
\end{equation}
This explicitly modulates the model's decoding distribution to align with the semantic requirements of $s_t$ (e.g., [Fast\_Step] triggers concise decoding), without requiring architectural modifications.
A full list of tags are provided in Appendix~\ref{app:tags}.

\begin{table*}[t]
\centering
\resizebox{\linewidth}{!}{
\begin{tabular}{c|c|cc|cc|cc|cc|cc}
\toprule[1.5pt]

\multirow{2}{*}{\textbf{Model}} & \multirow{2}{*}{\textbf{Method}} &
\multicolumn{2}{c}{\textbf{GSM8K}} &
\multicolumn{2}{c}{\textbf{MATH500}} &
\multicolumn{2}{c}{\textbf{AIME 2024}} &
\multicolumn{2}{c}{\textbf{AIME 2025}} &
\multicolumn{2}{c}{\textbf{AMC}} \\
& &
\textbf{Acc ($\uparrow$, \%)} & \textbf{Tokens ($\downarrow$)} &
\textbf{Acc ($\uparrow$, \%)} & \textbf{Tokens ($\downarrow$)} &
\textbf{Acc ($\uparrow$, \%)} & \textbf{Tokens ($\downarrow$)} &
\textbf{Acc ($\uparrow$, \%)} & \textbf{Tokens ($\downarrow$)} &
\textbf{Acc ($\uparrow$, \%)} & \textbf{Tokens ($\downarrow$)} \\
\midrule[1.5pt]

& COT & \textbf{95.1} & 2136 & 95.0 & 4823 & 60.0 & 12662 & 50.0 & 12720 & 92.5 & 7408 \\
& DEER & 94.5 (\textcolor{myyellow}{-0.6}) & 1250 (\textcolor{myblue}{-41\%}) &
        92.6 (\textcolor{myyellow}{-2.4}) & 3214 (\textcolor{myblue}{-33\%}) &
        \textbf{63.3 (\textcolor{myyellow}{+3.3})} & 9327 (\textcolor{myblue}{-26\%}) &
        \textbf{55.0 (\textcolor{myyellow}{+5.0})} & 12084 (\textcolor{myblue}{-5\%}) &
        87.5 (\textcolor{myyellow}{-5.0}) & 4906 (\textcolor{myblue}{-34\%}) \\
& CGRS & -- & -- &
        91.3 (\textcolor{myyellow}{-3.7}) & \textbf{2704 (\textcolor{myblue}{-44\%})} &
        56.7 (\textcolor{myyellow}{-3.3}) & \textbf{7893 (\textcolor{myblue}{-37\%})} &
        -- & -- &
        86.7 (\textcolor{myyellow}{-5.8}) & \textbf{4351 (\textcolor{myblue}{-41\%})} \\
\multirow{-4}{*}{\textbf{Qwen3-4B}} & \cellcolor{gray!10}\textbf{\method} &
\cellcolor{gray!10}\textbf{95.1 (\textcolor{myyellow}{+0.0})} & \cellcolor{gray!10}\textbf{845 (\textcolor{myblue}{-60\%})} &
\cellcolor{gray!10}\textbf{95.6 (\textcolor{myyellow}{+0.6})} & \cellcolor{gray!10}2833 (\textcolor{myblue}{-41\%}) &
\cellcolor{gray!10}60.0 (\textcolor{myyellow}{+0.0}) & \cellcolor{gray!10}9462 (\textcolor{myblue}{-25\%}) &
\cellcolor{gray!10}53.3 (\textcolor{myyellow}{+3.3}) & \cellcolor{gray!10}\textbf{9907 (\textcolor{myblue}{-22\%})} &
\cellcolor{gray!10}\textbf{92.5 (\textcolor{myyellow}{+0.0})} & \cellcolor{gray!10}5255 (\textcolor{myblue}{-29\%}) \\
\midrule

& COT & \textbf{95.8} & 2152 & 95.6 & 5166 & \textbf{66.7} & 12393 & \textbf{60.0} & 12835 & 90.0 & 7920 \\
& DEER & 95.5 (\textcolor{myyellow}{-0.3}) & 981 (\textcolor{myblue}{-54\%}) &
        92.6 (\textcolor{myyellow}{-3.0}) & \textbf{2732 (\textcolor{myblue}{-47\%})} &
        61.7 (\textcolor{myyellow}{-5.0}) & 8796 (\textcolor{myblue}{-29\%}) &
        \textbf{60.0 (\textcolor{myyellow}{+0.0})} & 12229 (\textcolor{myblue}{-5\%}) &
        \textbf{92.5 (\textcolor{myyellow}{+2.5})} & \textbf{4392 (\textcolor{myblue}{-45\%})} \\
& CGRS & -- & -- &
        93.3 (\textcolor{myyellow}{-2.3}) & 3507 (\textcolor{myblue}{-32\%}) &
        61.1 (\textcolor{myyellow}{-5.6}) & \textbf{8792 (\textcolor{myblue}{-29\%})} &
        -- & -- &
        89.2 (\textcolor{myyellow}{-0.8}) & 5595 (\textcolor{myblue}{-29\%}) \\
\multirow{-4}{*}{\textbf{Qwen3-8B}} & \cellcolor{gray!10}\textbf{\method} &
\cellcolor{gray!10}95.5 (\textcolor{myyellow}{-0.3}) & \cellcolor{gray!10}\textbf{879 (\textcolor{myblue}{-60\%})} &
\cellcolor{gray!10}\textbf{95.8 (\textcolor{myyellow}{+0.2})} & \cellcolor{gray!10}3215 (\textcolor{myblue}{-38\%}) &
\cellcolor{gray!10}\textbf{66.7 (\textcolor{myyellow}{+0.0})} & \cellcolor{gray!10}9674 (\textcolor{myblue}{-22\%}) &
\cellcolor{gray!10}\textbf{60.0 (\textcolor{myyellow}{+0.0})} & \cellcolor{gray!10}\textbf{10939 (\textcolor{myblue}{-15\%})} &
\cellcolor{gray!10}\textbf{92.5 (\textcolor{myyellow}{+2.5})} & \cellcolor{gray!10}5308 (\textcolor{myblue}{-33\%}) \\
\midrule

& COT & 95.7 & 1660 & 96.2 & 4598 & 73.3 & 11742 & 56.7 & 12820 & 92.5 & 6964 \\
& DEER & 95.3 (\textcolor{myyellow}{-0.4}) & 840 (\textcolor{myblue}{-49\%}) &
        94.0 (\textcolor{myyellow}{-2.2}) & 3074 (\textcolor{myblue}{-33\%}) &
        \textbf{76.7 (\textcolor{myyellow}{+3.4})} & \textbf{7619 (\textcolor{myblue}{-35\%})} &
        \textbf{66.7 (\textcolor{myyellow}{+10.0})} & 11135 (\textcolor{myblue}{-13\%}) &
        95.0 (\textcolor{myyellow}{+2.5}) & \textbf{4763 (\textcolor{myblue}{-32\%})} \\
& CGRS & -- & -- &
        94.5 (\textcolor{myyellow}{-1.7}) & 3235 (\textcolor{myblue}{-30\%}) &
        70.0 (\textcolor{myyellow}{-3.3}) & 8662 (\textcolor{myblue}{-26\%}) &
        -- & -- &
        93.3 (\textcolor{myyellow}{+0.8}) & 5076 (\textcolor{myblue}{-27\%}) \\
\multirow{-4}{*}{\textbf{Qwen3-14B}} & \cellcolor{gray!10}\textbf{\method} &
\cellcolor{gray!10}\textbf{96.4 (\textcolor{myyellow}{+0.7})} & \cellcolor{gray!10}\textbf{767 (\textcolor{myblue}{-54\%})} &
\cellcolor{gray!10}\textbf{96.6 (\textcolor{myyellow}{+0.4})} & \cellcolor{gray!10}\textbf{2904 (\textcolor{myblue}{-37\%})} &
\cellcolor{gray!10}73.3 (\textcolor{myyellow}{+0.0}) & \cellcolor{gray!10}9626 (\textcolor{myblue}{-18\%}) &
\cellcolor{gray!10}60.0 (\textcolor{myyellow}{+3.3}) & \cellcolor{gray!10}\textbf{10637 (\textcolor{myblue}{-17\%})} &
\cellcolor{gray!10}\textbf{100.0 (\textcolor{myyellow}{+7.5})} & \cellcolor{gray!10}4848 (\textcolor{myblue}{-30\%}) \\
\midrule

& COT & 96.0 & 1688 & 95.6 & 4358 & 70.0 & 10788 & 63.3 & 12203 & 95.0 & 6448 \\
& DEER & 96.2 (\textcolor{myyellow}{+0.2}) & 769 (\textcolor{myblue}{-54\%}) &
        94.2 (\textcolor{myyellow}{-1.4}) & 3418 (\textcolor{myblue}{-22\%}) &
        \textbf{76.7 (\textcolor{myyellow}{+6.7})} & 8682 (\textcolor{myblue}{-20\%}) &
        66.7 (\textcolor{myyellow}{+3.4}) & 10893 (\textcolor{myblue}{-11\%}) &
        \textbf{97.5 (\textcolor{myyellow}{+2.5})} & 5753 (\textcolor{myblue}{-11\%}) \\
& CGRS & -- & -- &
        93.1 (\textcolor{myyellow}{-2.5}) & 2993 (\textcolor{myblue}{-31\%}) &
        65.6 (\textcolor{myyellow}{-4.4}) & \textbf{8128 (\textcolor{myblue}{-25\%})} &
        -- & -- &
        94.2 (\textcolor{myyellow}{-0.8}) & \textbf{4766 (\textcolor{myblue}{-26\%})} \\
\multirow{-4}{*}{\textbf{Qwen3-32B}} & \cellcolor{gray!10}\textbf{\method} &
\cellcolor{gray!10}\textbf{96.5 (\textcolor{myyellow}{+0.5})} & \cellcolor{gray!10}\textbf{680 (\textcolor{myblue}{-60\%})} &
\cellcolor{gray!10}\textbf{96.6 (\textcolor{myyellow}{+1.0})} & \cellcolor{gray!10}\textbf{2719 (\textcolor{myblue}{-38\%})} &
\cellcolor{gray!10}70.0 (\textcolor{myyellow}{+0.0}) & \cellcolor{gray!10}9183 (\textcolor{myblue}{-15\%}) &
\cellcolor{gray!10}\textbf{73.3 (\textcolor{myyellow}{+10.0})} & \cellcolor{gray!10}\textbf{9839 (\textcolor{myblue}{-19\%})} &
\cellcolor{gray!10}\textbf{97.5 (\textcolor{myyellow}{+2.5})} & \cellcolor{gray!10}4953 (\textcolor{myblue}{-23\%}) \\
\midrule

& COT & 77.6 & 1193 & 82.2 & 4723 & \textbf{26.7} & 11941 & 23.3 & 11879 & 67.5 & 7729 \\
& DEER & 74.7 (\textcolor{myyellow}{-2.9}) & 984 (\textcolor{myblue}{-18\%}) &
        67.8 (\textcolor{myyellow}{-14.4}) & \textbf{2497 (\textcolor{myblue}{-47\%})} &
        23.3 (\textcolor{myyellow}{-3.4}) & 9553 (\textcolor{myblue}{-20\%}) &
        10.0 (\textcolor{myyellow}{-13.3}) & 9281 (\textcolor{myblue}{-22\%}) &
        60.0 (\textcolor{myyellow}{-7.5}) & 5496 (\textcolor{myblue}{-29\%}) \\
& ThinkSwitcher & \textbf{84.7} (\textcolor{myyellow}{+7.1}) & 2114 (\textcolor{myblue}{+77\%}) &
        \textbf{82.4 (\textcolor{myyellow}{+0.2})} & 4544 (\textcolor{myblue}{-4\%}) &
        23.3 (\textcolor{myyellow}{-3.4}) & \textbf{8192 (\textcolor{myblue}{-31\%})} &
        \textbf{28.3 (\textcolor{myyellow}{+5.0})} & \textbf{6689 (\textcolor{myblue}{-43\%})} &
        -- & -- \\
\multirow{-4}{*}{\textbf{DS-Qwen-1.5B}} & \cellcolor{gray!10}\textbf{\method} &
\cellcolor{gray!10}77.3 (\textcolor{myyellow}{-0.3}) & \cellcolor{gray!10}\textbf{564 (\textcolor{myblue}{-53\%})} &
\cellcolor{gray!10}\textbf{82.4 (\textcolor{myyellow}{+0.2})} & \cellcolor{gray!10}3230 (\textcolor{myblue}{-32\%}) &
\cellcolor{gray!10}23.3 (\textcolor{myyellow}{-3.4}) & \cellcolor{gray!10}11211 (\textcolor{myblue}{-6\%}) &
\cellcolor{gray!10}23.3 (\textcolor{myyellow}{+0.0}) & \cellcolor{gray!10}9691 (\textcolor{myblue}{-18\%}) &
\cellcolor{gray!10}\textbf{70.0 (\textcolor{myyellow}{+2.5})} & \cellcolor{gray!10}\textbf{6670 (\textcolor{myblue}{-14\%})} \\
\midrule

& COT & 89.9 & 532 & 92.8 & 3537 & 50.0 & 12662 & 30.0 & 11028 & 87.5 & 6366 \\
& DEER & 90.6 (\textcolor{myyellow}{+0.7}) & 917 (\textcolor{myblue}{+72\%}) &
        89.8 (\textcolor{myyellow}{-3.0}) & 2143 (\textcolor{myblue}{-39\%}) &
        49.2 (\textcolor{myyellow}{-0.8}) & 9839 (\textcolor{myblue}{-22\%}) &
        36.7 (\textcolor{myyellow}{+6.7}) & 7257 (\textcolor{myblue}{-34\%}) &
        85.0 (\textcolor{myyellow}{-2.5}) & 4451 (\textcolor{myblue}{-30\%}) \\
& CGRS & -- & -- &
        87.6 (\textcolor{myyellow}{-5.2}) & \textbf{1867 (\textcolor{myblue}{-47\%})} &
        52.2 (\textcolor{myyellow}{+2.2}) & \textbf{7597 (\textcolor{myblue}{-40\%})} &
        -- & -- &
        88.3 (\textcolor{myyellow}{+0.8}) & \textbf{3406 (\textcolor{myblue}{-46\%})} \\
& ThinkSwitcher & \textbf{92.5 (\textcolor{myyellow}{+2.6})} & 1389 (\textcolor{myblue}{+161\%}) &
    91.3 (\textcolor{myyellow}{-1.5}) & 3495 (\textcolor{myblue}{-1.0\%}) &
        48.3 (\textcolor{myyellow}{-1.7})& 7936 (\textcolor{myblue}{-37\%}) &
        \textbf{37.5 (\textcolor{myyellow}{+7.5})} & \textbf{6948 (\textcolor{myblue}{-37\%})} &
        -- & -- \\
\multirow{-5}{*}{\textbf{DS-Qwen-7B}} & \cellcolor{gray!10}\textbf{\method} &
\cellcolor{gray!10}89.3 (\textcolor{myyellow}{-0.6}) & \cellcolor{gray!10}\textbf{385 (\textcolor{myblue}{-28\%})} &
\cellcolor{gray!10}\textbf{92.8 (\textcolor{myyellow}{+0.0})} & \cellcolor{gray!10}2237 (\textcolor{myblue}{-37\%}) &
\cellcolor{gray!10}\textbf{56.7 (\textcolor{myyellow}{+6.7})} & \cellcolor{gray!10}7978 (\textcolor{myblue}{-37\%}) &
\cellcolor{gray!10}36.7 (\textcolor{myyellow}{+6.7}) & \cellcolor{gray!10}9715 (\textcolor{myblue}{-12\%}) &
\cellcolor{gray!10}\textbf{92.5 (\textcolor{myyellow}{+5.0})} & \cellcolor{gray!10}3770 (\textcolor{myblue}{-41\%}) \\
\midrule

& COT & 94.9 & 1122 & 94.4 & 3539 & 60.0 & 10343 & 36.7 & 11002 & \textbf{92.5} & 5333 \\
& DEER & 93.3 (\textcolor{myyellow}{-1.6}) & 1040 (\textcolor{myblue}{-7\%}) &
        89.8 (\textcolor{myyellow}{-4.6}) & 2577 (\textcolor{myblue}{-27\%}) &
        \textbf{68.4 (\textcolor{myyellow}{+8.4})} & 8115 (\textcolor{myblue}{-22\%}) &
        36.7 (\textcolor{myyellow}{+0.0}) & 10125 (\textcolor{myblue}{-8\%}) &
        85.0 (\textcolor{myyellow}{-7.5}) & \textbf{4240 (\textcolor{myblue}{-20\%})} \\
& ThinkSwitcher & 94.3 (\textcolor{myyellow}{-0.6}) & 1042 (\textcolor{myblue}{-7\%}) &
        92.7 (\textcolor{myyellow}{-1.7}) & 3572 (\textcolor{myblue}{+1\%}) &
        60.4 (\textcolor{myyellow}{+0.4}) & \textbf{8044 (\textcolor{myblue}{-22\%})} &
        42.5 (\textcolor{myyellow}{+5.8}) & 10065 (\textcolor{myblue}{-9\%}) &
        -- & -- \\
\multirow{-4}{*}{\textbf{DS-Qwen-14B}} & \cellcolor{gray!10}\textbf{\method} &
\cellcolor{gray!10}\textbf{95.2 (\textcolor{myyellow}{+0.3})} & \cellcolor{gray!10}\textbf{621 (\textcolor{myblue}{-45\%})} &
\cellcolor{gray!10}\textbf{94.8 (\textcolor{myyellow}{+0.4})} & \cellcolor{gray!10}\textbf{2515 (\textcolor{myblue}{-29\%})} &
\cellcolor{gray!10}60.0 (\textcolor{myyellow}{+0.0}) & \cellcolor{gray!10}8393 (\textcolor{myblue}{-19\%}) &
\cellcolor{gray!10}\textbf{50.0 (\textcolor{myyellow}{+13.3})} & \cellcolor{gray!10}\textbf{8901 (\textcolor{myblue}{-19\%})} &
\cellcolor{gray!10}\textbf{92.5 (\textcolor{myyellow}{+0.0})} & \cellcolor{gray!10}4656 (\textcolor{myblue}{-13\%}) \\
\midrule

& COT & \textbf{90.2} & 1400 & 93.8 & 3412 & 56.7 & 10280 & \textbf{40} & 10893 & 90 & 5997 \\

& DEER & 89.8 (\textcolor{myyellow}{-0.4}) & 1473 (\textcolor{myblue}{+5\%}) &
91.4 (\textcolor{myyellow}{-2.4}) & 2995 (\textcolor{myblue}{-12\%}) & \textbf{66.7 (\textcolor{myyellow}{+10})} & 9755 (\textcolor{myblue}{-5\%}) & 36.7 (\textcolor{myyellow}{-3.3}) & 11820 (\textcolor{myblue}{+9\%}) & 90 (\textcolor{myyellow}{+0.0}) & 5408 (\textcolor{myblue}{-10\%}) \\

\multirow{-3}{*}{\textbf{Nemo-Llama-8b}} & \cellcolor{gray!10}\textbf{\method} &
\cellcolor{gray!10}89.7 (\textcolor{myyellow}{-0.5}) & \cellcolor{gray!10}\textbf{1035 (\textcolor{myblue}{-26\%})} &
\cellcolor{gray!10}\textbf{94 (\textcolor{myyellow}{+0.2})} & \cellcolor{gray!10}\textbf{2844 (\textcolor{myblue}{-17\%})} &
\cellcolor{gray!10}60 (\textcolor{myyellow}{+3.3}) & \cellcolor{gray!10}\textbf{9258 (\textcolor{myblue}{-10\%})} &
\cellcolor{gray!10}\textbf{40 (\textcolor{myyellow}{+0.0})} & \cellcolor{gray!10}\textbf{10400 (\textcolor{myblue}{-5\%})} &
\cellcolor{gray!10}\textbf{95 (\textcolor{myyellow}{+5})} & \cellcolor{gray!10}\textbf{4441 (\textcolor{myblue}{-26\%})} \\
\midrule

& COT & \textbf{97.0} & 1561 & \textbf{97.0} & 4025 & 66.7 & 11305 & \textbf{63.3} & 12554 & 90.0 & 7086 \\
& DEER & 96.3 (\textcolor{myyellow}{-0.7}) & 977 (\textcolor{myblue}{-37\%}) &
        94.6 (\textcolor{myyellow}{-2.4}) & 3316 (\textcolor{myblue}{-18\%}) &
        \textbf{70.0 (\textcolor{myyellow}{+3.3})} & 10097 (\textcolor{myblue}{-11\%}) &
        50.0 (\textcolor{myyellow}{-13.3}) & 11598 (\textcolor{myblue}{-8\%}) &
        \textbf{95.0 (\textcolor{myyellow}{+5.0})} & 5782 (\textcolor{myblue}{-18\%}) \\
& CGRS & -- & -- &
        94.2 (\textcolor{myyellow}{-2.8}) & \textbf{2810 (\textcolor{myblue}{-30\%})} &
        68.9 (\textcolor{myyellow}{-1.1}) & \textbf{8202 (\textcolor{myblue}{-27\%})} &
        -- & -- &
        93.3 (\textcolor{myyellow}{+3.3}) & \textbf{4771 (\textcolor{myblue}{-33\%})} \\
\multirow{-4}{*}{\textbf{QwQ-32B}} & \cellcolor{gray!10}\textbf{\method} &
\cellcolor{gray!10}96.6 (\textcolor{myyellow}{-0.4}) & \cellcolor{gray!10}\textbf{969 (\textcolor{myblue}{-38\%})} &
\cellcolor{gray!10}\textbf{97.0 (\textcolor{myyellow}{+0.0})} & \cellcolor{gray!10}3256 (\textcolor{myblue}{-19\%}) &
\cellcolor{gray!10}\textbf{70.0 (\textcolor{myyellow}{+3.3})} & \cellcolor{gray!10}9181 (\textcolor{myblue}{-19\%}) &
\cellcolor{gray!10}53.3 (\textcolor{myyellow}{-10.0}) & \cellcolor{gray!10}\textbf{11416 (\textcolor{myblue}{-9\%})} &
\cellcolor{gray!10}\textbf{95.0 (\textcolor{myyellow}{+5.0})} & \cellcolor{gray!10}5777 (\textcolor{myblue}{-18\%}) \\
\bottomrule[1.5pt]
\end{tabular}
}
\caption{{Performance on mathematical reasoning benchmarks.}
Metrics include \textbf{Acc} ($\uparrow$) and \textbf{Tokens} ($\downarrow$). Changes relative to the COT are highlighted in \textcolor{myyellow}{orange} for Acc and \textcolor{myblue}{blue} for Tokens. Best results within each group are bolded.}
\label{tab:main_results}
\end{table*}

\begin{figure*}[!t]
    \centering
    \includegraphics[width=\linewidth]{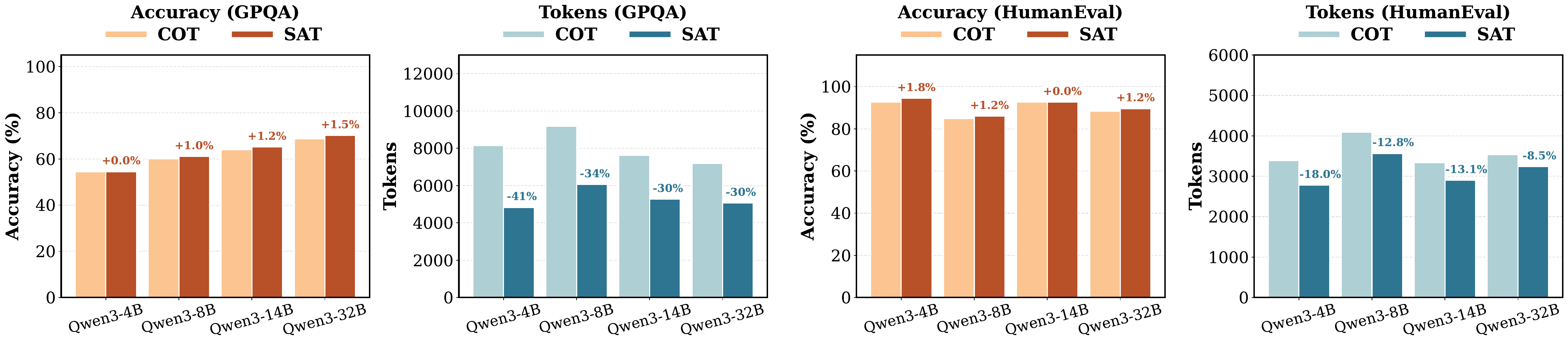}
    \vspace{-1.5em}
    \caption{Accuracy and token usage on {GPQA} and {HumanEval} across different model scales.}
    \label{fig:cross_domain}
\end{figure*}

\section{Experiment}
\subsection{Experimental Setup}
\label{sec:experimental_setup}

 \paragraph{Datasets and Evaluation Metrics.} 
To comprehensively evaluate \method's performance across diverse domains and difficulty levels, we conduct experiments on 7 benchmarks covering mathematical, scientific, and code reasoning. 
For \textbf{mathematical reasoning}, we evaluate general capabilities on GSM8K~\citep{gsm8k} and MATH 500~\citep{math500}, and competition-level performance on AMC 2023~\citep{amc2023}, AIME 2024~\citep{aime2024} and AIME 2025~\citep{aime2025}.
For \textbf{scientific reasoning}, we utilize the expert-level GPQA Diamond~\citep{gpqa}. 
For \textbf{code reasoning}, we adopt HumanEval~\citep{human_eval}. 
We report Accuracy (\textbf{Acc}, Pass@1) and average generated tokens per query (\textbf{Token}) as evaluation metrics.

\noindent\textbf{Backbone LRMs and Baselines.} 
We evaluate \method across diverse backbones: Qwen3 series~\citep{qwen3}, DeepSeek-R1-Distill-Qwen series~\citep{ds_qwen}, Llama-3.1-Nemotron-8B~\citep{Llama-Nemotron} and QwQ-32B~\citep{qwq32b}. 
We compare \method against three categories of baseline methods: 
(1) \textbf{COT}: Chain-of-thought reasoning; 
(2) \textbf{DEER}~\citep{deer}: the SoTA early-exit method based on confidence truncation;
(3) \textbf{CGRS}~\citep{huang2026efficient}: the SoTA suppression method that suppresses the sampling of reflection-triggering tokens.
(4) \textbf{ThinkSwitcher}~\citep{thinkswitcher}: a recent \emph{question-driven routing} approach that uses a switcher to select between short and long COT.
Due to space constraints, comparisons with additional methods are detailed in Appendix~\ref{app:more_baselines}.

\noindent\textbf{Implementation Details.} 
All our methods are implemented based on the \textit{HuggingFace Transformers} framework and conducted on NVIDIA H20 (96GB) GPUs. We set sampling parameters to $\text{Temperature}=0.6$, $\text{Top-}p=0.95$ and a max length of 16,384 tokens.
For the FSM, we set the difficulty thresholds to $\tau_{\texttt{fast}}=0.2$, $\tau_{\texttt{slow}}=0.6$. The hysteresis margin is set to $\Delta=0.1$. Details are provided in Appendix~\ref{app:implementation}.

\subsection{Main Results}
\label{sec:main_results}

\paragraph{Overall Performance.}
As summarized in Table~\ref{tab:main_results}, \method demonstrates a generally superior accuracy–efficiency balance compared to Vanilla COT and other baselines across five mathematical reasoning benchmarks.\footnote{
See Appendix~\ref{app:more_baselines} for more baseline comparisons.}
On average, over all tested models and datasets, \method reduces token usage by 25.1\% while improving accuracy by +1.5 points.
In comparison, while DEER truncates more aggressively—reducing tokens by 47\% on MATH‑500 (Qwen3‑8B)—it often degrades accuracy (e.g., –3.0 points on the same setting).
Similarly, CGRS can introduce notable performance drops (e.g., –2.3 accuracy on MATH‑500 for Qwen3‑8B).
ThinkSwitcher, as a question-driven routing baseline, can reduce computation on some datasets (e.g., \textbf{-43\%} Tok on AIME 2025 for DS-Qwen-1.5B) but its gains are less consistent across tasks and models, underscoring the advantage of \method’s stepwise online modulation.

\noindent\textbf{Generalization across Domains and Model Scales.}
As illustrated in Figure~\ref{fig:cross_domain}, \method maintains strong generalization across both scientific reasoning (GPQA Diamond) and code reasoning (HumanEval) tasks, achieving consistent efficiency improvements with nearly no compromise in accuracy.
Moreover, \method scales stably across models of varying sizes, achieving an average reduction of 33.8\% in tokens on GPQA with a +0.75\% gain in accuracy, and 13.1\% fewer tokens on HumanEval alongside a +1.05\% improvement in accuracy.

\subsection{Ablation Study}
\label{sec:ablation}

\begin{figure}[t]
    \centering
    \includegraphics[width=\linewidth]{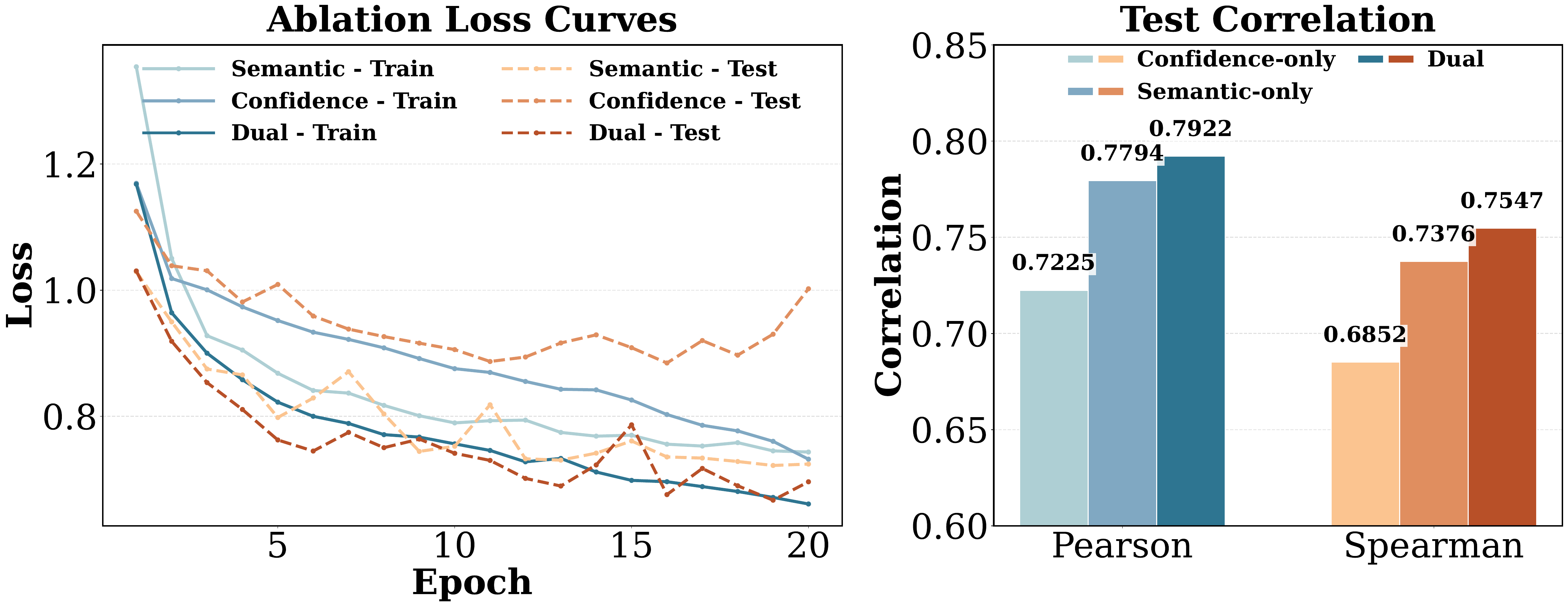}
    \vspace{-0.4cm}
    \caption{{Ablation study results.} \textit{Left:} training and test loss curves for GRU-based Pilot under different inputs. \textit{Right:} test-set correlation (Pearson/Spearman) between predicted step scores and supervision targets.}
    \label{fig:ablation_two_panels}
\end{figure}

\noindent\textbf{Pilot Ablation: dual vs.\ single-source features.}
Figure~\ref{fig:ablation_two_panels} compares three variants of the Pilot module: \textit{confidence-only} (relying solely on uncertainty features), \textit{semantic-only} (using only GTE-small embeddings), and the \textit{dual-input} model.
Overall, the dual-input model consistently shows the best optimization behavior and test performance. 
While semantic features outperform confidence features alone, their combination yields better performance than either feature source alone.
These results indicate that confidence features—capturing local uncertainty dynamics—and semantic features—encoding step content—provide complementary signals for difficulty estimation.


\begin{figure}[t]
    \centering
    \includegraphics[width=\linewidth]{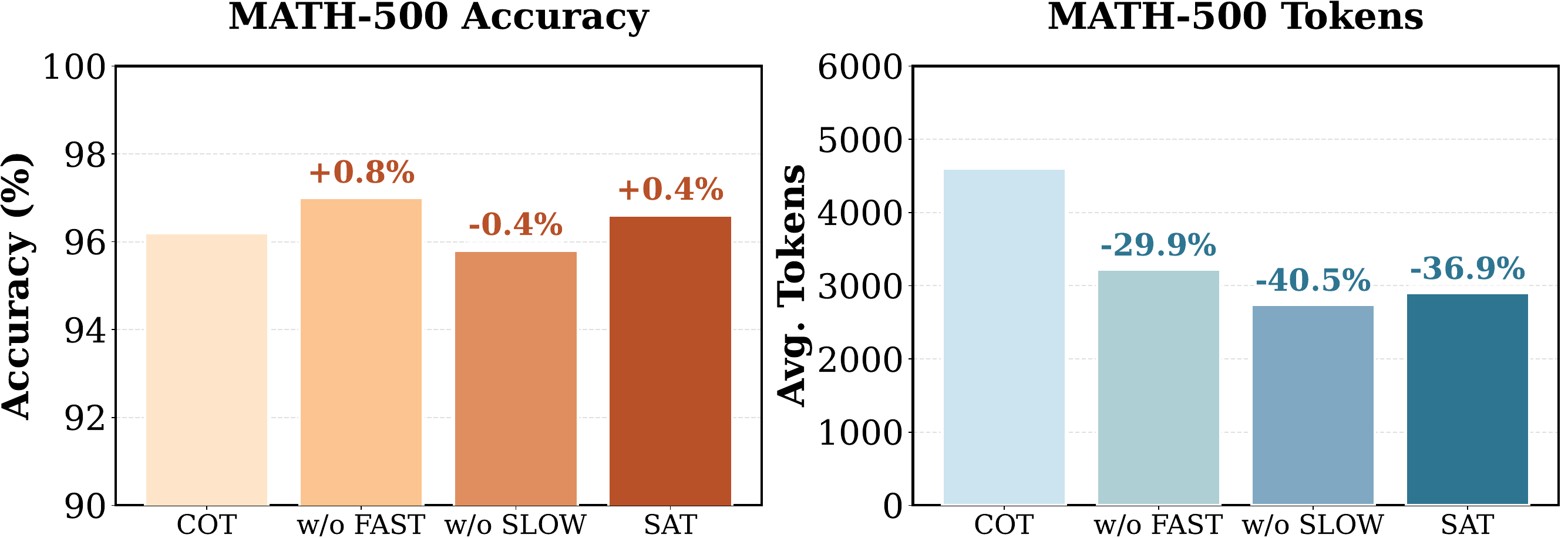}
    \caption{{Strategy ablation results on MATH500 (Qwen3-14B).} Accuracy and token usage of \textit{COT}, \textit{w/o FAST}, \textit{w/o SLOW}, and \method.}
    \label{fig:math500_ablation_results}
    \vspace{-0.3cm}
\end{figure}

\noindent\textbf{Strategy Ablation: \textsc{Fast} vs.\ \textsc{Slow}.}
Figure~\ref{fig:math500_ablation_results} presents an ablation study on MATH500 (Qwen3-14B) to assess the contribution of each thinking mode.
Removing the \textsc{Fast} mode (\textit{w/o FAST}) yields a slight accuracy gain (+0.8\% over COT), whereas disabling \textsc{Slow} (\textit{w/o SLOW}) leads to a small decline (-0.4\%), indicating that \textsc{Slow} reasoning is essential for reasoning.
Meanwhile, \method preserves accuracy while saving tokens (+0.4\% on accuracy; -36.9\% on token usage).
All ablated variants reduce token usage, with \textit{w/o FAST} saving 29.9\% and \textit{w/o SLOW} achieving the largest compression (40.5\%, 2737).
Interestingly, \textit{w/o FAST} also reduces token cost; Appendix~\ref{app:exp_process_behavior} further shows it decreases reflective behaviors, and we hypothesize this comes from fewer ``error--repair--re-verify'' loops.

\subsection{Analysis}
\label{sec:discussion}


\begin{figure}[t]
    \centering
    \includegraphics[width=\linewidth]{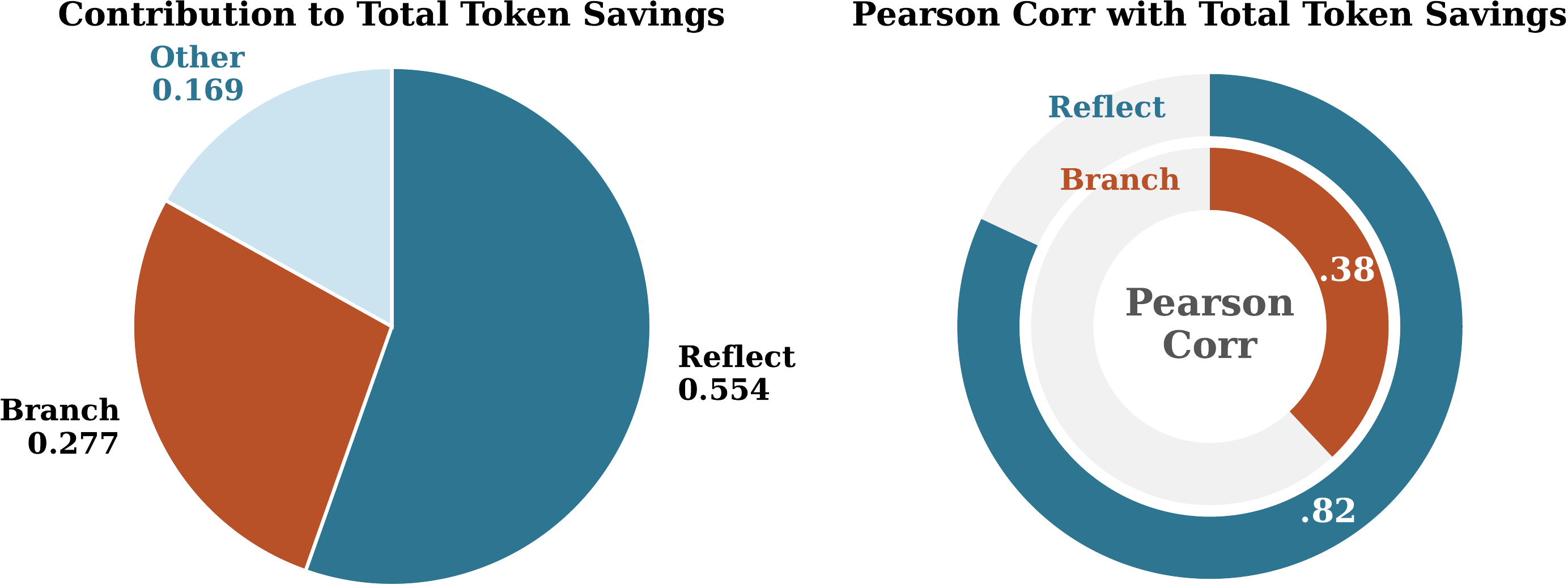}
    \caption{{Attribution analysis of token savings on MATH500 (Qwen3-14B).} \textit{Left:} decomposition of \method's total token savings into reflection-related and branching-related components. \textit{Right:} Pearson correlation between per-sample total token savings and each component.}
    \label{fig:token_saving_attribution}
    \vspace{-0.3cm}
\end{figure}

\paragraph{Why can \method save tokens?}
Analyzing Qwen3-14B on MATH500 (Figure~\ref{fig:token_saving_attribution}), \method's efficiency stems mainly from reduced reflection (contribution 0.554, Pearson $r{=}0.818$) and secondarily from limited branching (contribution 0.277, Pearson $r{=}0.374$). This confirms that curbing redundant reflection is the primary driver of token savings, with branching control providing auxiliary benefits (See details settings in Appendix~\ref{app:why_short_window}).

\begin{figure}[t]
    \centering
    \includegraphics[width=0.95\linewidth]{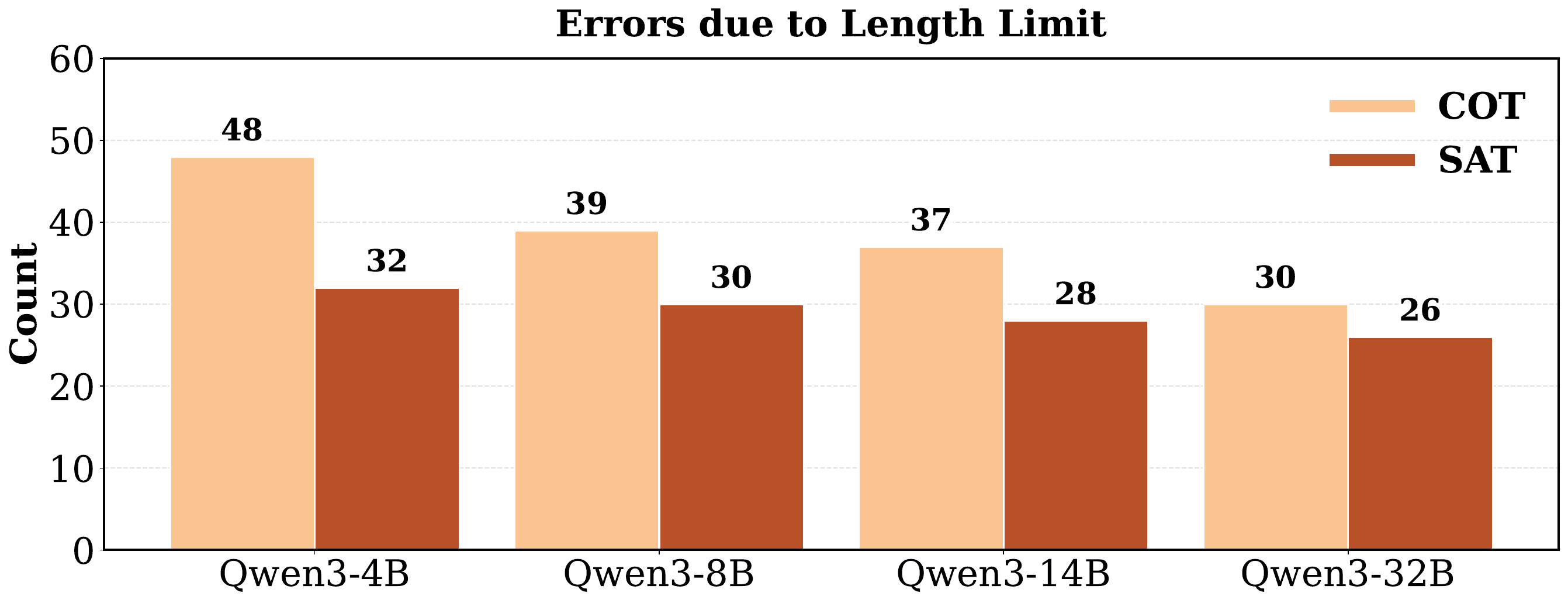}  
    \vspace{-0.2cm}
    \caption{{Length-limit failures on aggregated math benchmarks.}
    Number of incorrect predictions caused by exceeding maximum length, aggregated over GSM8K, MATH500, AIME 2024, AIME 2025, and AMC.}
    \label{fig:len_limit_errors_sum}
    \vspace{-0.4cm}
\end{figure}

\begin{figure}[t]
    \centering
    \includegraphics[width=\linewidth]{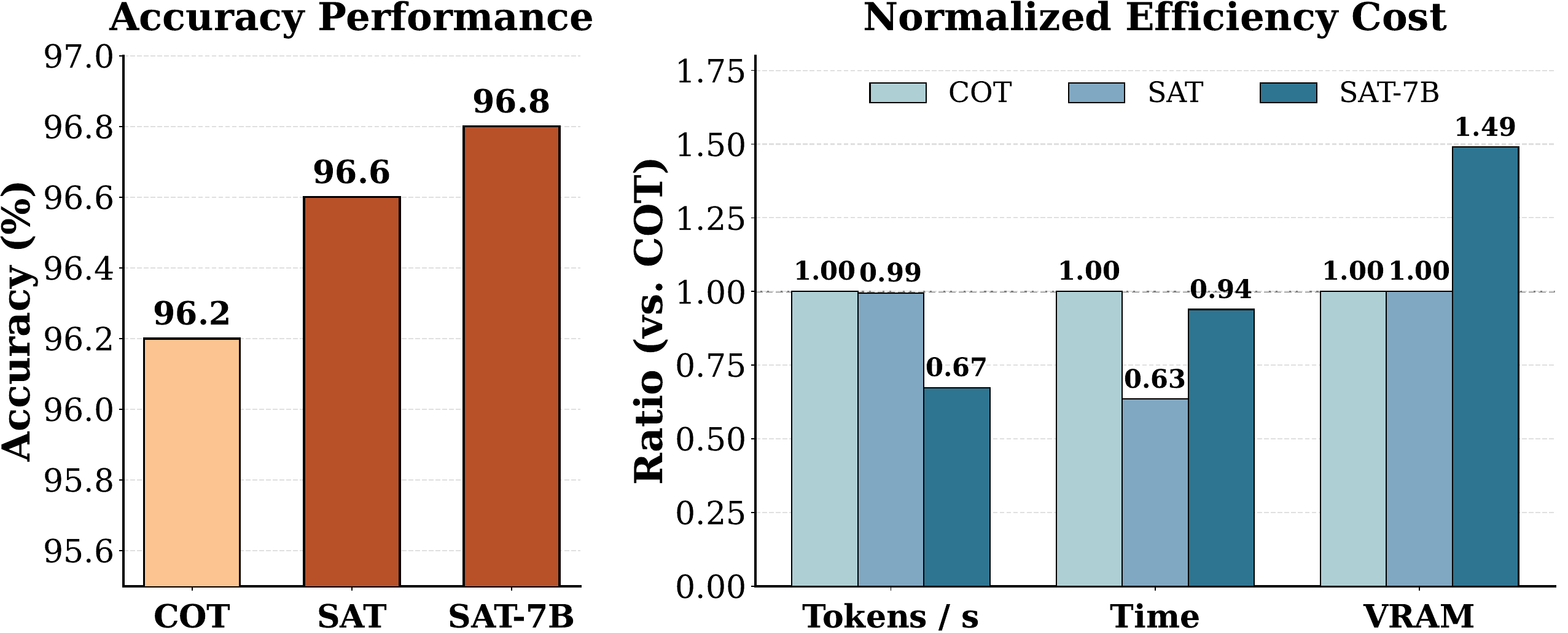} 
    \caption{
    {Performance analysis on MATH500 (Qwen3-14B).}
    {Left:} Absolute accuracy comparison.
    {Right:} Normalized computational costs (Tokens/s, Time, VRAM) relative to the COT baseline ($1.0\times$).
}
    \label{fig:efficiency_analysis}
\end{figure}

\noindent\textbf{Why can \method slightly improve accuracy?} Despite its efficiency focus, \method achieves consistent accuracy gains by optimizing generation budgets. 
By pruning redundant reasoning, it prevents valid solutions from exceeding the context limit (16k). As shown in Figure~\ref{fig:len_limit_errors_sum}, \method reduces length-limit failures in the Qwen3 model series from \textbf{154} to \textbf{116} (\textbf{-24.7\%}), effectively salvaging samples that would otherwise fail due to premature truncation. Moreover, by curbing error–repair–re-verify loops, \method reduces stochastic drift from repeated self-corrections, yielding additional accuracy gains.


\noindent\textbf{Additional Computation Cost.}
We assess the computational overhead of \method by comparing it with standard COT and a variant using a 7B-scale pilot model (SAT‑7B). 
As shown in Figure~\ref{fig:efficiency_analysis} (left), \method retains COT's throughput (29.98 vs. 30.14 tokens/s) while matching the accuracy ($\sim$96.6\%) of the heavier baselines, validating the efficacy of the 30M PRM. 
Crucially, Figure~\ref{fig:efficiency_analysis} (right) reveals that while SAT-7B's overhead ($1.5\times$ VRAM) negates its savings, \method maintains COT-level memory ($\sim$30GB) and translates token reduction into a $\sim$37\% end-to-end speedup. 
This confirms \method achieves practical acceleration with negligible overhead (details in Appendix~\ref{sec:appendix_throughput}).

\begin{figure}[t]
    \centering
    \includegraphics[width=0.95\linewidth]{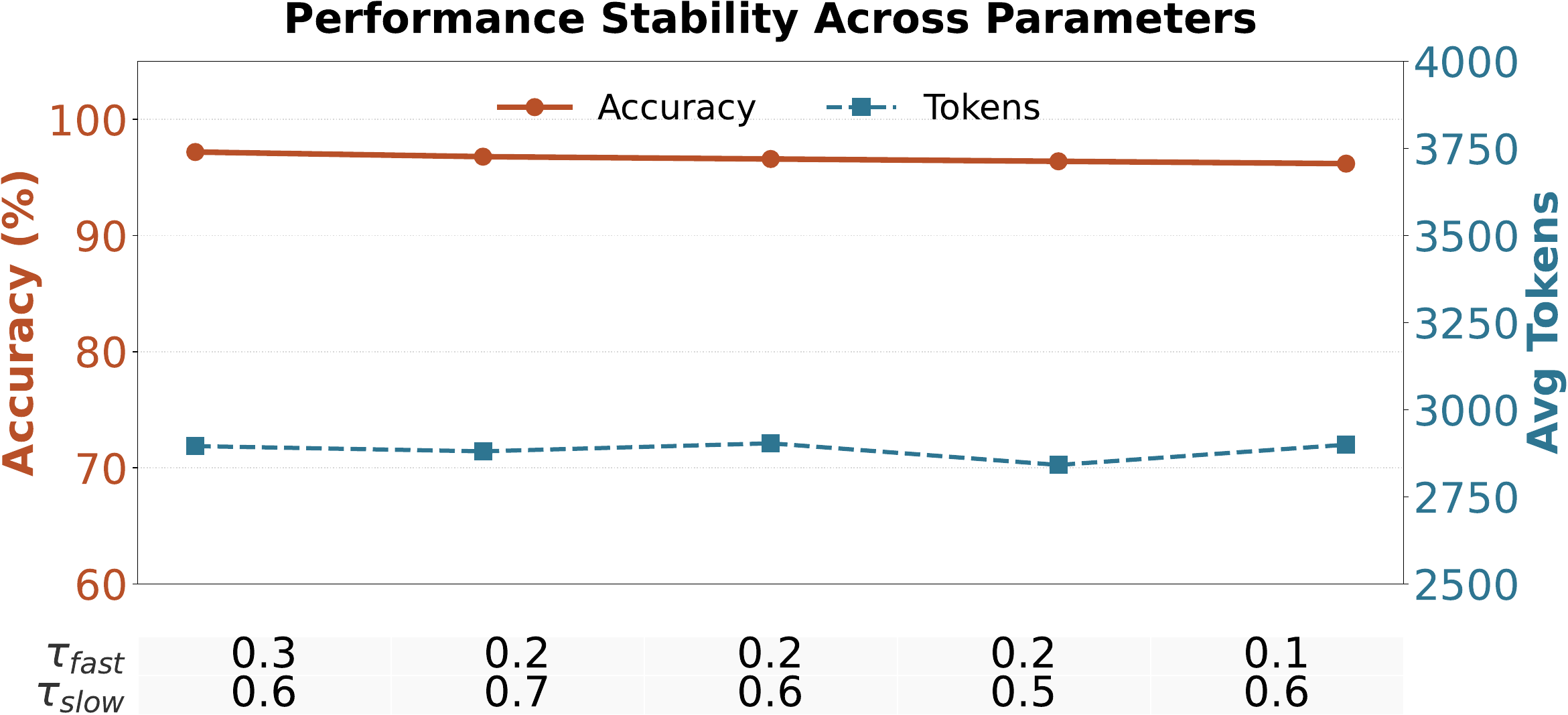}
    \caption{{Hyperparameter robustness on MATH500 (Qwen3-14B).} Sensitivity of \method to $\tau_{\texttt{fast}}$ and $\tau_{\texttt{slow}}$.}
    \label{fig:param_robust}
    \vspace{-0.5cm}
\end{figure}

\noindent\textbf{Hyperparameter Robustness.}
We further analyze the choice of hyperparameter on  MATH500 using Qwen3-14B.
Figure~\ref{fig:param_robust} shows that \method is highly robust to threshold choices: under mild perturbations of $(\tau_{\texttt{fast}},\tau_{\texttt{slow}})$ (e.g., $\tau_{\texttt{fast}}\!\in\!\{0.3,0.2,0.1\}$ and $\tau_{\texttt{slow}}\!\in\!\{0.7,0.6,0.5\}$), accuracy remains consistently high (i.e., $\approx$96.6\%$\pm$0.6), while average tokens stay in a narrow band (2842--2904; $\approx$2.1\% relative range), indicating stable accuracy--efficiency trade-offs across settings.

\begin{figure}[t]
    \centering
    \includegraphics[width=\linewidth]{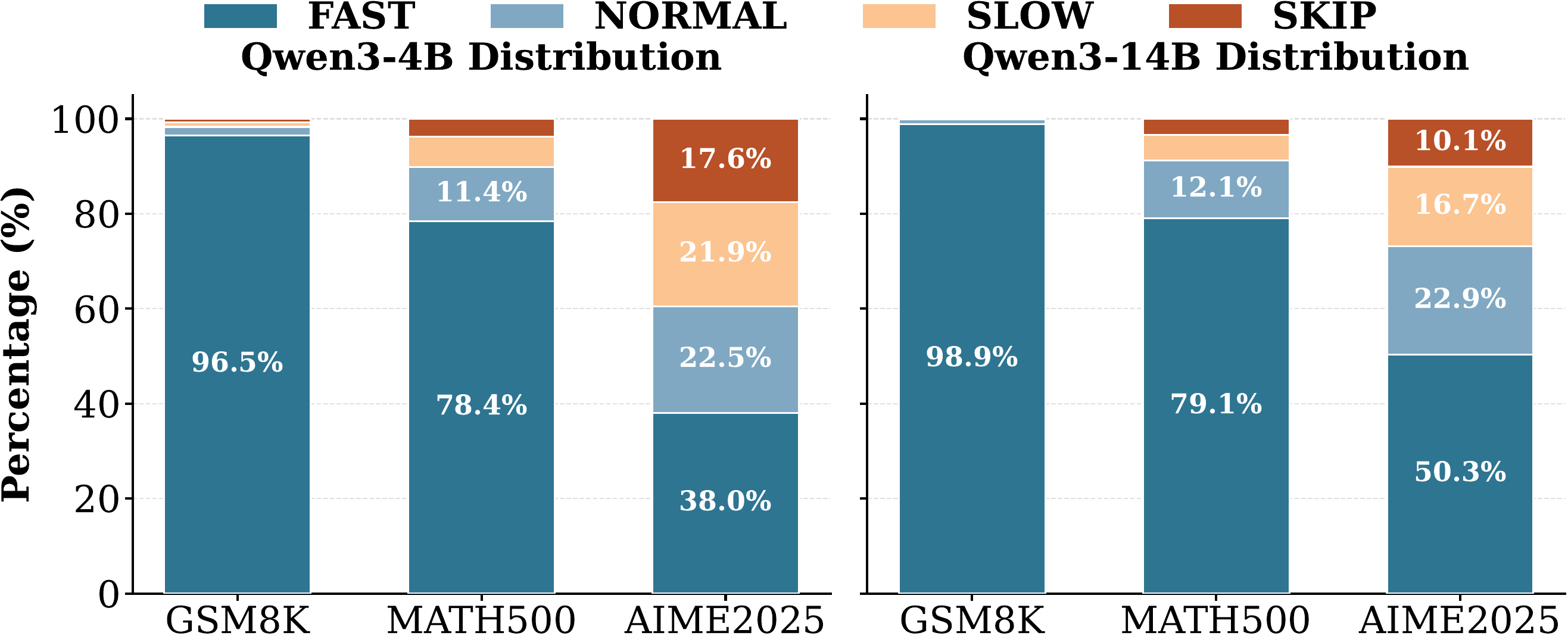}
    \caption{
    {State allocation across dataset difficulty and model scales.}
    We report the step-level thinking mode ratios on {GSM8K} (easy), {MATH500} (medium), and {AIME 2025} (hard), for {Qwen3-14B} and {Qwen3-4B}.
    }
    \label{fig:qwen3_mode_ratio}
\end{figure}

\noindent\textbf{Difficulty- and Scale-aware State Allocation.}
Figure~\ref{fig:qwen3_mode_ratio} depicts how \method dynamically shifts thinking modes in response to problem difficulty and model scale.
While dominating GSM8K with \textsc{Fast} (\textbf{98.9\%}), \method progressively reallocates compute for challenging tasks: on AIME 2025, \textsc{Fast} drops to \textbf{50.3\%} while \textsc{Slow}+\textsc{Skip} rises to \textbf{26.8\%} (Qwen3-14B). 
Crucially, the behavior is also sensitive to model capability: the weaker Qwen3-4B triggers significantly more deep reasoning on AIME 2025 than 14B (\textbf{39.5\%} vs.\ \textbf{26.8\%}).

\begin{figure}[t]
    \centering
    \includegraphics[width=0.95\linewidth]{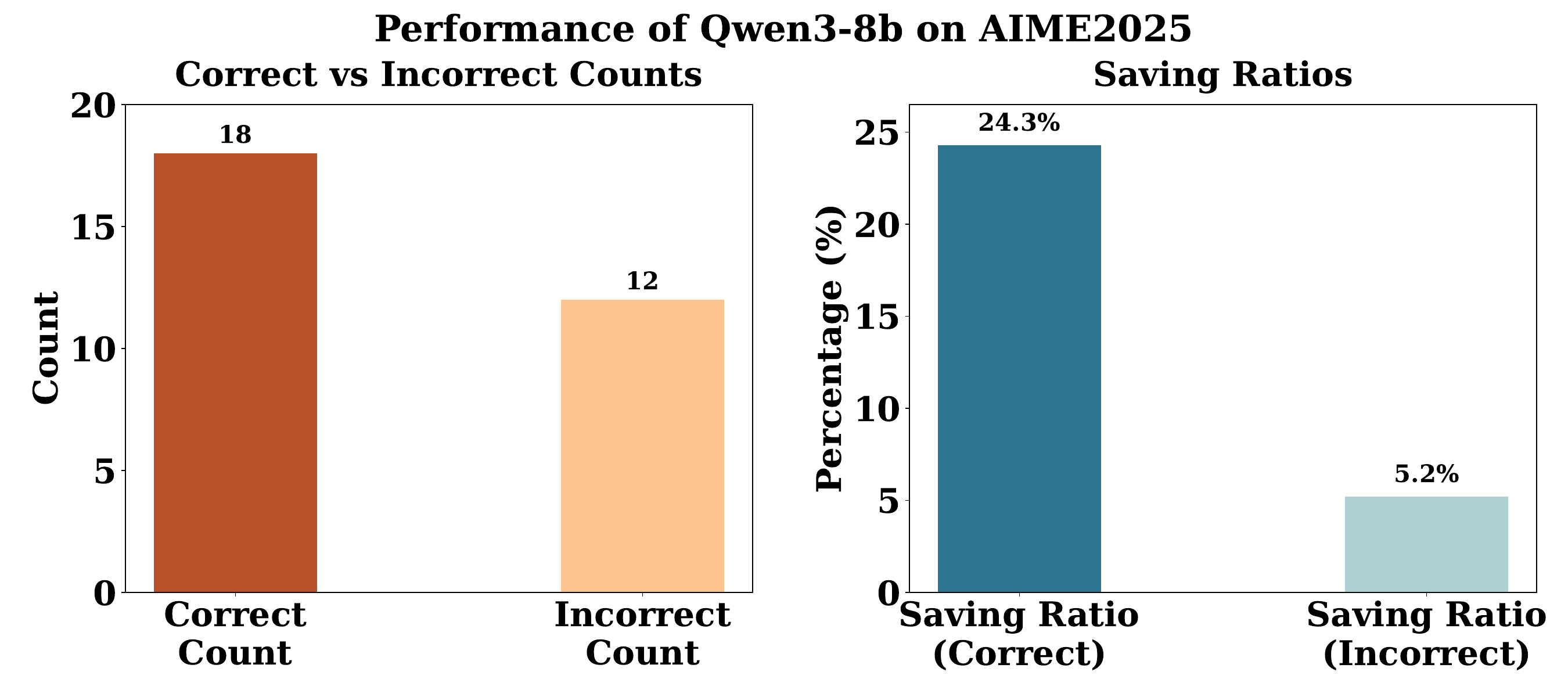}
    \vspace{-0.2cm}
    \caption{{Outcome-conditioned token savings on AIME2025 (Qwen3-8B).} SAT saves substantially more tokens on correct instances than on failed ones, where many runs hit the max-token limit.}
    \label{fig:difficulty_redundancy_bold}
    \vspace{-0.3cm}
\end{figure}

\noindent\textbf{Efficiency Gains are Outcome-Sensitive.}
Figure~\ref{fig:difficulty_redundancy_bold} reveals that savings are outcome-dependent: \method saves \textbf{24.3\%} on correct instances but only \textbf{5.2\%} on incorrect ones (AIME 2025). 
This aligns with the intuition that hard problems demand extensive exploration, where intermediate detours are necessary rather than redundant. 
Notably, most failures (10/12) hit the max generation limit, indicating active search saturation that leaves minimal headroom for safe pruning on extreme difficulties.

\section{Conclusion}
We propose \method, a framework that dynamically modulates reasoning depth via a lightweight, stepwise difficulty estimator. By navigating a Finite-State Machine, \method decouples reasoning efficacy from computational cost, achieving substantial efficiency gains (up to 40\% token reduction) without compromising accuracy. Our work validates that precise, step-level intervention is a viable and superior alternative to coarse-grained routing or static prompting for efficient reasoning deployment.

\section*{Limitations}

We identify two limitations in our current framework.
First, while the Pilot is designed to be lightweight, integrating an external module introduces a marginal computational overhead during the perception phase, although this is largely offset by the efficiency gains from token pruning.
Second, our method leverages in-context steering to modulate reasoning depth. Consequently, the precise execution of the state-machine logic relies on the backbone model's inherent instruction-following capabilities.


\bibliography{custom}
\appendix

\begin{table*}[t]
\centering
\small
\resizebox{\linewidth}{!}{
\begin{tabular}{l|l|p{8cm}}
\toprule[1.5pt]
\textbf{Category} & \textbf{Feature Name} & \textbf{Description \& Definition} \\
\midrule
\multirow{2}{*}{\textbf{Semantic}} 
& Step Embedding & The 384-dimensional dense vector encoded by \texttt{gte-small} representing the semantic content of the current reasoning step text. \\
\midrule
\multirow{12}{*}{\shortstack{\textbf{Uncertainty}\\(Static)}} 
& \texttt{canonical\_logprobs} & The log-probability of the sampled token: $\log P(x_t \mid x_{<t})$. \\
& \texttt{canonical\_selected\_rank} & The rank of the sampled token in the vocabulary distribution (1-based). \\
& \texttt{canonical\_entropy} & The Shannon entropy of the local probability distribution (truncated to Top-$K$). \\
& \texttt{canonical\_logit\_gap} & The difference between the largest logit (Top-1) and the second largest logit (Top-2). \\
& \texttt{canonical\_margin} & The probability margin between the Top-1 and Top-2 tokens: $P(x_{\text{top1}}) - P(x_{\text{top2}})$. \\
& \texttt{canonical\_topk\_mass@5} & The cumulative probability mass of the top-5 tokens. \\
& \texttt{canonical\_topk\_mass@10} & The cumulative probability mass of the top-10 tokens. \\
\midrule
\multirow{5}{*}{\shortstack{\textbf{Uncertainty}\\(Dynamic)}} 
& \texttt{canonical\_d\_logp} & First-order difference of log-probabilities: $\texttt{logp}_t - \texttt{logp}_{t-1}$. \\
& \texttt{canonical\_d\_entropy} & First-order difference of entropy: $\texttt{ent}_t - \texttt{ent}_{t-1}$. \\
& \texttt{canonical\_d\_margin} & First-order difference of margin: $\texttt{margin}_t - \texttt{margin}_{t-1}$. \\
& \texttt{canonical\_z\_logp} & Sliding-window Z-score of log-probability, capturing local anomalies relative to a history window (window size $W_z=20$). \\
\bottomrule[1.5pt]
\end{tabular}
}
\caption{{List of Input Features for the Lightweight Pilot.} The features are categorized into Semantic inputs and Uncertainty inputs (subdivided into Static and Dynamic metrics).}
\label{tab:feature_list}
\end{table*}

\begin{table*}[t]
    \centering
    
    \small
    \renewcommand{\arraystretch}{1.3} 
    \begin{tabular}{l|p{12cm}}
        \toprule[1.5pt]
        \textbf{Control Tag} & \textbf{Instruction / Semantics} \\
        \midrule
        \texttt{[Fast\_Step]} & Indicates the current step appears easy. The model is instructed to keep reasoning brief and avoid unnecessary details. \\
        \midrule
        \texttt{[Slow\_Step]} & Indicates the current step appears difficult. The model is instructed to perform detailed, expansive reasoning. \\
        \midrule
        \texttt{[Normal\_Step]} & Indicates moderate difficulty. The model is instructed to resume standard step-by-step reasoning. (Corresponds to the \textsc{Normal} state). \\
        \midrule
        \texttt{[Skip\_Step]} & Indicates the step is excessively difficult and further elaboration is unlikely to be helpful. The model is instructed to summarize existing reasoning, make a reasonable guess, and quickly output the final answer. \\
        \bottomrule[1.5pt]
    \end{tabular}
    \caption{{Definitions of Control Tags.} These tags are inserted into the context to modulate the reasoning style of the subsequent step.}
    \label{tab:control_tags}
\end{table*}

\section{Extended Motivation and Case Study}
\label{app:intro_motivation}

\begin{figure}[h]
    \centering
    \includegraphics[width=\linewidth]{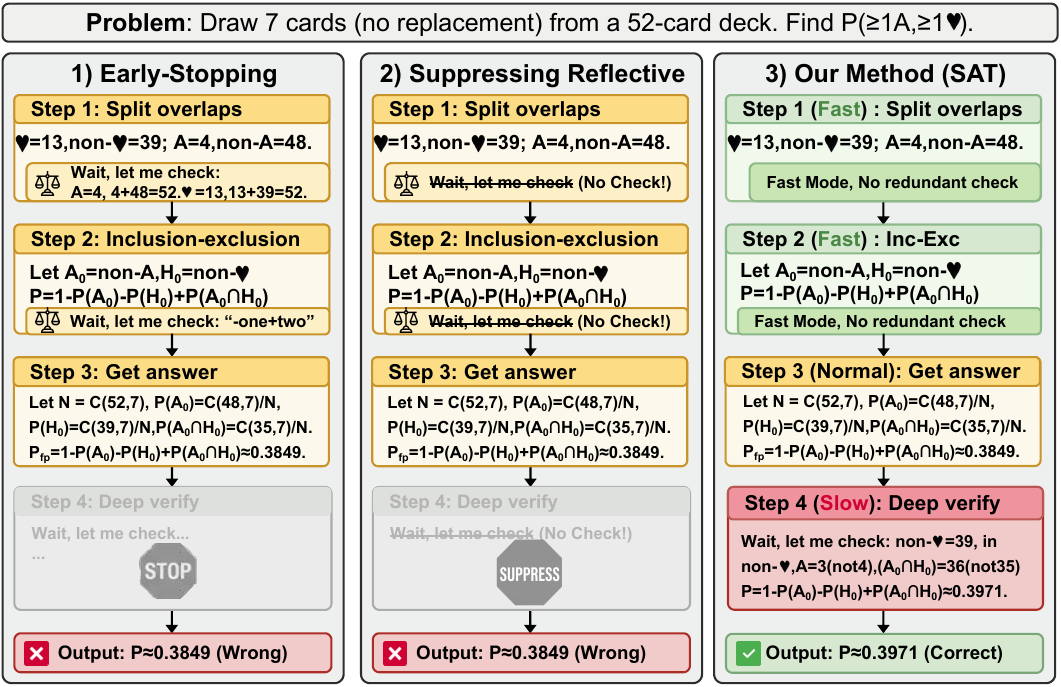}
    \caption{\textbf{Baselines vs. SAT.} 
    Early-Stopping halts after a high-confidence first-pass answer and fails; Suppressing Reflective eliminates all verification steps, preventing necessary self-correction and leading to the same error. In contrast, \method skips redundancy on easy steps but preserves verification on hard steps, achieving the correct answer efficiently.}
\label{fig:motivation_appendix}
\end{figure}

In this section, we elaborate on the motivation behind \method by analyzing a specific failure mode in complex reasoning tasks, as illustrated in Figure~\ref{fig:motivation_appendix}.

The example task involves a probability problem requiring the inclusion-exclusion principle: \textit{"Draw 7 cards from a 52-card deck. Find $P(\ge 1A, \ge 1\heartsuit)$."} As shown in Figure~\ref{fig:motivation_appendix}, we compare three approaches:

\begin{itemize}
    \item \textbf{Early-Stopping (Baseline 1):} The model wastes compute performing redundant checks on trivial steps (e.g., verifying basic arithmetic like $13+39=52$ in Step 1). Crucially, it exits the reasoning process too early. After deriving the first-pass answer ($P_{fp} \approx 0.3849$), the model halts without performing a deep verification of the intersection term $P(A_0 \cap H_0)$, leading to an incorrect result.
    
    \item \textbf{Suppressing Reflective (Baseline 2):} This method aggressively removes all "checking" tokens (e.g., "Wait, let me check...") to maximize speed. While this reduces token usage, it removes the model's ability to self-correct. Consequently, it executes the standard, flawed logic for the inclusion-exclusion step and arrives at the same incorrect probability ($0.3849$) as the Early-Stopping baseline.
    
    \item \textbf{Our Method (SAT):} \method demonstrates dynamic compute allocation. 
    \begin{enumerate}
        \item \textbf{Fast Mode:} For Step 1 (Split overlaps) and Step 2 (Inclusion-Exclusion setup), the model identifies the sub-steps as "easy" and skips redundant checks, mimicking the speed of the suppression method.
        \item \textbf{Slow Mode:} In Step 4, the model detects the complexity of the verification required for the intersection term. It engages a "Deep verify" process, realizing that the number of Aces in the "Non-Heart" set is 3, not 4 (since the Ace of Hearts is excluded). 
    \end{enumerate}
    This targeted verification allows \method to correct the logic error and derive the true answer ($P \approx 0.3971$), balancing the efficiency of suppression with the rigor of deep reasoning.
\end{itemize}

\section{Details of Input Features}
\label{app:input_features}

As described in Section~\ref{subsec:difficulty}, our lightweight Pilot utilizes a dual-channel input representation $\mathbf{z}_t = [\mathbf{h}_{\text{unc}}; \mathbf{h}_{\text{sem}}]$.
The Semantic Features ($\mathbf{h}_{\text{sem}}$) are dense vector representations of the reasoning step's textual content, while the Uncertainty Features ($\mathbf{h}_{\text{unc}}$) are statistical metrics derived from the LLM's decoding logits.
Table~\ref{tab:feature_list} provides a detailed categorization of these features based on our implementation.

\paragraph{Feature Extraction Process.}
For every token generated within a reasoning step, we extract a set of raw confidence metrics (e.g., LogProb, Entropy) and dynamic metrics (e.g., Z-score, First-order differences).
Specifically, for a reasoning step $y_t$ consisting of multiple tokens, we aggregate these token-level metrics (via mean-pooling) to form the step-level feature vector $\mathbf{h}_{\text{unc}}$.
The semantic vector $\mathbf{h}_{\text{sem}}$ is obtained by encoding the raw text of the step using the frozen \texttt{gte-small} encoder.

\section{Control Tags and System Prompts}
\label{app:tags}

To enable dynamic control over the reasoning depth, we inject specific control tags into the model's context based on the state determined by the FSM. These tags are defined in the system prompt to guide the model's behavior via instruction following.

\paragraph{Control Tags.} 
Table~\ref{tab:control_tags} details the semantics of each tag used in our framework. Note that while the FSM defines a \textsc{Normal} state, we map this to the tag \texttt{[Normal\_Step]} in the prompt to explicitly signal a return to moderate reasoning depth.

\paragraph{System Prompt.} 
The full system prompt used to initialize the model and define these behaviors is provided below. This prompt is prepended to the user query. For code-generation benchmarks (e.g., HumanEval), we use a separate system prompt that requests the model to output only the final Python implementation within \texttt{```python} and \texttt{```} code blocks, to avoid format interference from \texttt{\textbackslash boxed\{\}}.

\begin{tcolorbox}[colback=gray!5!white,colframe=gray!75!black,title=System Prompt for In-Context Steering]
\small
\texttt{Please reason step by step, and put your final answer within \textbackslash boxed\{\}.}

\vspace{0.5em}
\texttt{During your thinking, you may see the following tags:}

\begin{itemize}[leftmargin=1.5em, noitemsep, topsep=0pt]
    \item \texttt{[Fast\_Step] means the current step seems easy; keep your reasoning brief and avoid unnecessary details.}
    \item \texttt{[Slow\_Step] means the current step seems difficult; please perform detailed reasoning.}
    \item \texttt{[Normal\_Step] means the current step has moderate difficulty; please resume normal step-by-step reasoning.}
    \item \texttt{[Skip\_Step] means this step is too difficult and further detailed expansion is not very helpful, please summarize the existing reasoning, make a reasonable guess for the conclusion, and then quickly output the final answer.}
\end{itemize}

\vspace{0.5em}
\texttt{\{question\}}
\end{tcolorbox}

\section{Implementation Details}
\label{app:implementation}

\subsection{Experimental Environment}
All experiments are conducted on \textbf{NVIDIA H20 (96GB)} GPUs using the \textbf{HuggingFace Transformers} library. 
Following official recommendations, we set sampling parameters to $\text{Temperature}=0.6$ and $\text{Top-}p=0.95$, with a maximum generation length limit of 16,384 tokens.

\subsection{Hyperparameter Settings}
For \method, the detailed FSM configurations are:
\begin{itemize}
    \item \textbf{Thresholds:} $\tau_{\texttt{fast}}=0.2$, $\tau_{\texttt{slow}}=0.6$, $\tau_{\texttt{skip}}=0.85$.
    \item \textbf{Hysteresis Margin:} $\Delta=0.1$ (to prevent state flickering).
    \item \textbf{History Windows:} $k_{\texttt{fast}}=6$, $k_{\texttt{slow}}=5$, $k_{\texttt{skip}}=35$.
\end{itemize}

\subsection{Baseline Reproduction}
For baselines DEER, CGRS, and ThinkSwitcher, we prioritize citing reported data from their original papers, while results for NoThinking, TALE, and Dynasor are sourced from the CGRS paper~\citep{huang2026efficient}.

\section{Additional Experimental Analysis}
\label{sec:appendix_exp}

\subsection{Comparisons with Additional Baselines}
\label{app:more_baselines}

Due to space limitations in the main text, we present comparisons with three additional baseline strategies in this section to provide a broader context for \method's performance. These methods represent distinct approaches to the efficiency-accuracy trade-off:

\begin{itemize}[leftmargin=1.5em]
    \item \textbf{NoThinking}: A direct prompting baseline that instructs the model to bypass intermediate reasoning processes entirely and generate the final answer directly. This method prioritizes maximum efficiency by eliminating the Chain-of-Thought.
    
    \item \textbf{TALE}~\citep{tale}: A constraint-based prompting strategy that explicitly instructs the model to solve the problem within a pre-defined token budget. By imposing a strict length constraint on the generation process, TALE encourages the model to condense its reasoning steps and prioritize essential information, thereby reducing computational costs while attempting to maintain reasoning integrity.
    
    \item \textbf{Dynasor}~\citep{fu2025efficiently}: A decoding manipulation method that periodically extracts intermediate answers at fixed token intervals during generation. It employs an early-exit mechanism, terminating the inference process if multiple consecutive checks yield consistent answers, thereby reducing redundant computation.
\end{itemize}

The comparative results are detailed in Table~\ref{tab:more_baselines_results} (below).

\begin{table*}[t]
\centering
\resizebox{\linewidth}{!}{
\begin{tabular}{c|c|cc|cc|cc|cc|cc}
\toprule[1.5pt]

\multirow{2}{*}{\textbf{Model}} & \multirow{2}{*}{\textbf{Method}} &
\multicolumn{2}{c}{\textbf{GSM8K}} &
\multicolumn{2}{c}{\textbf{MATH500}} &
\multicolumn{2}{c}{\textbf{AIME 2024}} &
\multicolumn{2}{c}{\textbf{AIME 2025}} &
\multicolumn{2}{c}{\textbf{AMC}} \\
& &
\textbf{Acc ($\uparrow$, \%)} & \textbf{Tokens ($\downarrow$)} &
\textbf{Acc ($\uparrow$, \%)} & \textbf{Tokens ($\downarrow$)} &
\textbf{Acc ($\uparrow$, \%)} & \textbf{Tokens ($\downarrow$)} &
\textbf{Acc ($\uparrow$, \%)} & \textbf{Tokens ($\downarrow$)} &
\textbf{Acc ($\uparrow$, \%)} & \textbf{Tokens ($\downarrow$)} \\
\midrule[1.5pt]

\multirow{-7}{*} 
& COT & \textbf{95.1} & 2136 & 95.0 & 4823 & 60.0 & 12662 & 50.0 & 12720 & 92.5 & 7408 \\
& TALE & -- & -- & 
    89.1 (\textcolor{myyellow}{-5.9}) & 2657 (\textcolor{myblue}{-45\%}) & 
    48.9 (\textcolor{myyellow}{-11.1}) & 9727 (\textcolor{myblue}{-23\%}) & 
    -- & -- & 
    86.7 (\textcolor{myyellow}{-5.8}) & 5107 (\textcolor{myblue}{-31\%}) \\
& NoThinking & -- & -- & 
    84.9 (\textcolor{myyellow}{-10.1}) & \textbf{988 (\textcolor{myblue}{-80\%})} & 
    24.4 (\textcolor{myyellow}{-35.6}) & \textbf{4504 (\textcolor{myblue}{-64\%})} & 
    -- & -- & 
    70.0 (\textcolor{myyellow}{-22.5}) & \textbf{1710 (\textcolor{myblue}{-77\%})} \\
& Dynasor & -- & -- & 
    90.1 (\textcolor{myyellow}{-4.9}) & 3877 (\textcolor{myblue}{-20\%}) & 
    54.3 (\textcolor{myyellow}{-5.7}) & 9912 (\textcolor{myblue}{-22\%}) & 
    -- & -- & 
    86.7 (\textcolor{myyellow}{-5.8}) & 6233 (\textcolor{myblue}{-16\%}) \\
& DEER & 94.5 (\textcolor{myyellow}{-0.6}) & 1250 (\textcolor{myblue}{-41\%}) &
    92.6 (\textcolor{myyellow}{-2.4}) & 3214 (\textcolor{myblue}{-33\%}) &
    \textbf{63.3 (\textcolor{myyellow}{+3.3})} & 9327 (\textcolor{myblue}{-26\%}) &
    \textbf{55.0 (\textcolor{myyellow}{+5.0})} & 12039 (\textcolor{myblue}{-5\%}) &
    87.5 (\textcolor{myyellow}{-5.0}) & 4906 (\textcolor{myblue}{-34\%}) \\
& CGRS & -- & -- &
    91.3 (\textcolor{myyellow}{-3.7}) & 2704 (\textcolor{myblue}{-44\%}) &
    56.7 (\textcolor{myyellow}{-3.3}) & 7893 (\textcolor{myblue}{-38\%}) &
    -- & -- &
    86.7 (\textcolor{myyellow}{-5.8}) & 4351 (\textcolor{myblue}{-41\%}) \\
\multirow{-7}{*}{\textbf{Qwen3-4B}} 
& \cellcolor{gray!10}\textbf{\method} &
    \cellcolor{gray!10}\textbf{95.1 (\textcolor{myyellow}{+0.0})} & \cellcolor{gray!10}\textbf{845 (\textcolor{myblue}{-60\%})} &
    \cellcolor{gray!10}\textbf{95.6 (\textcolor{myyellow}{+0.6})} & \cellcolor{gray!10}2833 (\textcolor{myblue}{-41\%}) &
    \cellcolor{gray!10}60.0 (\textcolor{myyellow}{+0.0}) & \cellcolor{gray!10}9462 (\textcolor{myblue}{-25\%}) &
    \cellcolor{gray!10}53.3 (\textcolor{myyellow}{+3.3}) & \cellcolor{gray!10}\textbf{9907 (\textcolor{myblue}{-22\%})} &
    \cellcolor{gray!10}\textbf{92.5 (\textcolor{myyellow}{+0.0})} & \cellcolor{gray!10}5255 (\textcolor{myblue}{-29\%}) \\
\midrule

& COT & \textbf{95.8} & 2152 & 95.6 & 5166 & 66.7 & 12393 & \textbf{60.0} & 12835 & 90.0 & 7920 \\
& TALE & -- & -- & 
    92.3 (\textcolor{myyellow}{-3.3}) & 3885 (\textcolor{myblue}{-25\%}) & 
    \textbf{68.9 (\textcolor{myyellow}{+2.2})} & 10942 (\textcolor{myblue}{-12\%}) & 
    -- & -- & 
    88.3 (\textcolor{myyellow}{-1.7}) & 6872 (\textcolor{myblue}{-13\%}) \\
& NoThinking & -- & -- & 
    87.1 (\textcolor{myyellow}{-8.5}) & \textbf{1239 (\textcolor{myblue}{-76\%})} & 
    30.0 (\textcolor{myyellow}{-36.7}) & \textbf{5967 (\textcolor{myblue}{-52\%})} & 
    -- & -- & 
    72.5 (\textcolor{myyellow}{-17.5}) & \textbf{2426 (\textcolor{myblue}{-69\%})} \\
& Dynasor & -- & -- & 
    91.7 (\textcolor{myyellow}{-3.9}) & 3841 (\textcolor{myblue}{-26\%}) & 
    62.2 (\textcolor{myyellow}{-4.5}) & 10174 (\textcolor{myblue}{-18\%}) & 
    -- & -- & 
    89.2 (\textcolor{myyellow}{-0.8}) & 6457 (\textcolor{myblue}{-18\%}) \\
& DEER & 95.5 (\textcolor{myyellow}{-0.3}) & 981 (\textcolor{myblue}{-54\%}) &
    92.6 (\textcolor{myyellow}{-3.0}) & 2732 (\textcolor{myblue}{-47\%}) &
    61.7 (\textcolor{myyellow}{-5.0}) & 8796 (\textcolor{myblue}{-29\%}) &
    \textbf{60.0 (\textcolor{myyellow}{+0.0})} & 12229 (\textcolor{myblue}{-5\%}) &
    \textbf{92.5 (\textcolor{myyellow}{+2.5})} & 4392 (\textcolor{myblue}{-45\%}) \\
& CGRS & -- & -- &
    93.3 (\textcolor{myyellow}{-2.3}) & 3507 (\textcolor{myblue}{-32\%}) &
    61.1 (\textcolor{myyellow}{-5.6}) & 8792 (\textcolor{myblue}{-29\%}) &
    -- & -- &
    89.2 (\textcolor{myyellow}{-0.8}) & 5595 (\textcolor{myblue}{-29\%}) \\
\multirow{-7}{*}{\textbf{Qwen3-8B}} & \cellcolor{gray!10}\textbf{\method} &
\cellcolor{gray!10}95.5 (\textcolor{myyellow}{-0.3}) & \cellcolor{gray!10}\textbf{879 (\textcolor{myblue}{-59\%})} &
\cellcolor{gray!10}\textbf{95.8 (\textcolor{myyellow}{+0.2})} & \cellcolor{gray!10}3215 (\textcolor{myblue}{-38\%}) &
\cellcolor{gray!10}66.7 (\textcolor{myyellow}{+0.0}) & \cellcolor{gray!10}9674 (\textcolor{myblue}{-22\%}) &
\cellcolor{gray!10}\textbf{60.0 (\textcolor{myyellow}{+0.0})} & \cellcolor{gray!10}\textbf{10939 (\textcolor{myblue}{-15\%})} &
\cellcolor{gray!10}\textbf{92.5 (\textcolor{myyellow}{+2.5})} & \cellcolor{gray!10}5308 (\textcolor{myblue}{-33\%}) \\
\midrule

& COT & 95.7 & 1660 & 96.2 & 4598 & 73.3 & 11742 & 56.7 & 12820 & 92.5 & 6964 \\
& TALE & -- & -- &
  93.7 (\textcolor{myyellow}{-2.5}) & 3389 (\textcolor{myblue}{-26\%}) &
  71.1 (\textcolor{myyellow}{-2.2}) & 10860 (\textcolor{myblue}{-8\%}) &
  -- & -- &
  92.5 (\textcolor{myyellow}{+0.0}) & 5951 (\textcolor{myblue}{-15\%}) \\
& NoThinking & -- & -- &
  87.0 (\textcolor{myyellow}{-9.2}) & \textbf{853 (\textcolor{myblue}{-81\%})} &
  27.8 (\textcolor{myyellow}{-45.5}) & \textbf{3689 (\textcolor{myblue}{-69\%})} &
  -- & -- &
  77.5 (\textcolor{myyellow}{-15.0}) & 1616 (\textcolor{myblue}{-77\%}) \\
& Dynasor & -- & -- &
  84.4 (\textcolor{myyellow}{-11.8}) & 3667 (\textcolor{myblue}{-20\%}) &
  65.6 (\textcolor{myyellow}{-7.7}) & 9775 (\textcolor{myblue}{-17\%}) &
  -- & -- &
  90.0 (\textcolor{myyellow}{-2.5}) & 6030 (\textcolor{myblue}{-13\%}) \\
& DEER & 95.3 (\textcolor{myyellow}{-0.4}) & 840 (\textcolor{myblue}{-49\%}) &
  94.0 (\textcolor{myyellow}{-2.2}) & 3074 (\textcolor{myblue}{-33\%}) &
  \textbf{76.7 (\textcolor{myyellow}{+3.4})} & 7619 (\textcolor{myblue}{-35\%}) &
  \textbf{66.7 (\textcolor{myyellow}{+10.0})} & 11135 (\textcolor{myblue}{-13\%}) &
  95.0 (\textcolor{myyellow}{+2.5}) & \textbf{4763 (\textcolor{myblue}{-32\%})} \\
& CGRS & -- & -- &
  94.5 (\textcolor{myyellow}{-1.7}) & 3235 (\textcolor{myblue}{-30\%}) &
  70.0 (\textcolor{myyellow}{-3.3}) & 8662 (\textcolor{myblue}{-26\%}) &
  -- & -- &
  93.3 (\textcolor{myyellow}{+0.8}) & 5076 (\textcolor{myblue}{-27\%}) \\
\multirow{-7}{*}{\textbf{Qwen3-14B}} & \cellcolor{gray!10}\textbf{\method} &
\cellcolor{gray!10}\textbf{96.4 (\textcolor{myyellow}{+0.7})} & \cellcolor{gray!10}\textbf{767 (\textcolor{myblue}{-54\%})} &
\cellcolor{gray!10}\textbf{96.6 (\textcolor{myyellow}{+0.4})} & \cellcolor{gray!10}2904 (\textcolor{myblue}{-37\%}) &
\cellcolor{gray!10}73.3 (\textcolor{myyellow}{+0.0}) & \cellcolor{gray!10}9626 (\textcolor{myblue}{-18\%}) &
\cellcolor{gray!10}60.0 (\textcolor{myyellow}{+3.3}) & \cellcolor{gray!10}\textbf{10637 (\textcolor{myblue}{-17\%})} &
\cellcolor{gray!10}\textbf{100.0 (\textcolor{myyellow}{+7.5})} & \cellcolor{gray!10}4848 (\textcolor{myblue}{-30\%}) \\
\midrule

& COT & 96.0 & 1688 & 95.6 & 4358 & 70.0 & 10788 & 63.3 & 12203 & 95.0 & 6448 \\
& TALE & -- & -- &
  93.6 (\textcolor{myyellow}{-2.0}) & 3857 (\textcolor{myblue}{-11\%}) &
  67.8 (\textcolor{myyellow}{-2.2}) & 10688 (\textcolor{myblue}{-1\%}) &
  -- & -- &
  93.3 (\textcolor{myyellow}{-1.7}) & 6533 (+1\%) \\
& NoThinking & -- & -- &
  87.0 (\textcolor{myyellow}{-8.6}) & \textbf{1054 (\textcolor{myblue}{-76\%})} &
  41.1 (\textcolor{myyellow}{-28.9}) & \textbf{5635 (\textcolor{myblue}{-48\%})} &
  -- & -- &
  75.0 (\textcolor{myyellow}{-20.0}) & \textbf{2221 (\textcolor{myblue}{-66\%})} \\
& Dynasor & -- & -- &
  85.2 (\textcolor{myyellow}{-10.4}) & 3486 (\textcolor{myblue}{-20\%}) &
  64.4 (\textcolor{myyellow}{-5.6}) & 9518 (\textcolor{myblue}{-12\%}) &
  -- & -- &
  92.5 (\textcolor{myyellow}{-2.5}) & 5521 (\textcolor{myblue}{-14\%}) \\
& DEER & 96.2 (\textcolor{myyellow}{+0.2}) & 769 (\textcolor{myblue}{-54\%}) &
  94.2 (\textcolor{myyellow}{-1.4}) & 3418 (\textcolor{myblue}{-22\%}) &
  \textbf{76.7 (\textcolor{myyellow}{+6.7})} & 8682 (\textcolor{myblue}{-20\%}) &
  66.7 (\textcolor{myyellow}{+3.4}) & 10893 (\textcolor{myblue}{-11\%}) &
  \textbf{97.5 (\textcolor{myyellow}{+2.5})} & 5753 (\textcolor{myblue}{-11\%}) \\
& CGRS & -- & -- &
  93.1 (\textcolor{myyellow}{-2.5}) & 2993 (\textcolor{myblue}{-31\%}) &
  65.6 (\textcolor{myyellow}{-4.4}) & \textbf{8128 (\textcolor{myblue}{-25\%})} &
  -- & -- &
  94.2 (\textcolor{myyellow}{-0.8}) & \textbf{4766 (\textcolor{myblue}{-26\%})} \\
\multirow{-7}{*}{\textbf{Qwen3-32B}} & \cellcolor{gray!10}\textbf{\method} &
\cellcolor{gray!10}\textbf{96.5 (\textcolor{myyellow}{+0.5})} & \cellcolor{gray!10}\textbf{680 (\textcolor{myblue}{-60\%})} &
\cellcolor{gray!10}\textbf{96.6 (\textcolor{myyellow}{+1.0})} & \cellcolor{gray!10}\textbf{2719 (\textcolor{myblue}{-38\%})} &
\cellcolor{gray!10}70.0 (\textcolor{myyellow}{+0.0}) & \cellcolor{gray!10}9183 (\textcolor{myblue}{-15\%}) &
\cellcolor{gray!10}\textbf{73.3 (\textcolor{myyellow}{+10.0})} & \cellcolor{gray!10}\textbf{9839 (\textcolor{myblue}{-19\%})} &
\cellcolor{gray!10}\textbf{97.5 (\textcolor{myyellow}{+2.5})} & \cellcolor{gray!10}4953 (\textcolor{myblue}{-23\%}) \\
\midrule

& COT & 77.6 & 1193 & 82.2 & 4723 & \textbf{26.7} & 11941 & 23.3 & 11879 & 67.5 & 7729 \\
& DEER & 74.7 (\textcolor{myyellow}{-2.9}) & 984 (\textcolor{myblue}{-18\%}) &
        67.8 (\textcolor{myyellow}{-14.4}) & \textbf{2497 (\textcolor{myblue}{-47\%})} &
        23.3 (\textcolor{myyellow}{-3.4}) & 9553 (\textcolor{myblue}{-20\%}) &
        10.0 (\textcolor{myyellow}{-13.3}) & 9281 (\textcolor{myblue}{-22\%}) &
        60.0 (\textcolor{myyellow}{-7.5}) & 5496 (\textcolor{myblue}{-29\%}) \\
& ThinkSwitcher & \textbf{84.7} (\textcolor{myyellow}{+7.1}) & 2114 (\textcolor{myblue}{+77\%}) &
        \textbf{82.4 (\textcolor{myyellow}{+0.2})} & 4544 (\textcolor{myblue}{-4\%}) &
        23.3 (\textcolor{myyellow}{-3.4}) & \textbf{8192 (\textcolor{myblue}{-31\%})} &
        \textbf{28.3 (\textcolor{myyellow}{+5.0})} & \textbf{6689 (\textcolor{myblue}{-43\%})} &
        -- & -- \\
\multirow{-4}{*}{\textbf{DS-Qwen-1.5B}} & \cellcolor{gray!10}\textbf{\method} &
\cellcolor{gray!10}77.3 (\textcolor{myyellow}{-0.3}) & \cellcolor{gray!10}\textbf{564 (\textcolor{myblue}{-53\%})} &
\cellcolor{gray!10}\textbf{82.4 (\textcolor{myyellow}{+0.2})} & \cellcolor{gray!10}3230 (\textcolor{myblue}{-32\%}) &
\cellcolor{gray!10}23.3 (\textcolor{myyellow}{-3.4}) & \cellcolor{gray!10}11211 (\textcolor{myblue}{-6\%}) &
\cellcolor{gray!10}23.3 (\textcolor{myyellow}{+0.0}) & \cellcolor{gray!10}9691 (\textcolor{myblue}{-18\%}) &
\cellcolor{gray!10}\textbf{70.0 (\textcolor{myyellow}{+2.5})} & \cellcolor{gray!10}\textbf{6670 (\textcolor{myblue}{-14\%})} \\
\midrule
& COT & 89.9 & 532 & 92.8 & 3537 & 50.0 & 12662 & 30.0 & 11028 & 87.5 & 6366 \\
& TALE & -- & -- &
  89.1 (\textcolor{myyellow}{-3.7}) & 2657 (\textcolor{myblue}{-25\%}) &
  48.9 (\textcolor{myyellow}{-1.1}) & 9727 (\textcolor{myblue}{-23\%}) &
  -- & -- &
  86.7 (\textcolor{myyellow}{-0.8}) & 5107 (\textcolor{myblue}{-20\%}) \\
& NoThinking & -- & -- &
  80.9 (\textcolor{myyellow}{-11.9}) & \textbf{1173 (\textcolor{myblue}{-67\%})} &
  32.2 (\textcolor{myyellow}{-17.8}) & \textbf{6680 (\textcolor{myblue}{-47\%})} &
  -- & -- &
  75.8 (\textcolor{myyellow}{-11.7}) & \textbf{2499 (\textcolor{myblue}{-61\%})} \\
& Dynasor & -- & -- &
  81.8 (\textcolor{myyellow}{-11.0}) & 2070 (\textcolor{myblue}{-41\%}) &
  47.8 (\textcolor{myyellow}{-2.2}) & 8334 (\textcolor{myblue}{-34\%}) &
  -- & -- &
  84.2 (\textcolor{myyellow}{-3.3}) & 5201 (\textcolor{myblue}{-18\%}) \\
& DEER & 90.6 (\textcolor{myyellow}{+0.7}) & 917 (\textcolor{myblue}{+72\%}) &
  89.8 (\textcolor{myyellow}{-3.0}) & 2143 (\textcolor{myblue}{-39\%}) &
  49.2 (\textcolor{myyellow}{-0.8}) & 9839 (\textcolor{myblue}{-22\%}) &
  36.7 (\textcolor{myyellow}{+6.7}) & 7257 (\textcolor{myblue}{-34\%}) &
  85.0 (\textcolor{myyellow}{-2.5}) & 4451 (\textcolor{myblue}{-30\%}) \\
& CGRS & -- & -- &
  87.6 (\textcolor{myyellow}{-5.2}) & \textbf{1867 (\textcolor{myblue}{-47\%})} &
  52.2 (\textcolor{myyellow}{+2.2}) & \textbf{7597 (\textcolor{myblue}{-40\%})} &
  -- & -- &
  88.3 (\textcolor{myyellow}{+0.8}) & \textbf{3406 (\textcolor{myblue}{-46\%})} \\
& ThinkSwitcher & \textbf{92.5 (\textcolor{myyellow}{+2.6})} & 1389 (\textcolor{myblue}{+161\%}) &
  91.3 (\textcolor{myyellow}{-1.5}) & 3495 (\textcolor{myblue}{-1.0\%}) &
  48.3 (\textcolor{myyellow}{-1.7})& 7936 (\textcolor{myblue}{-37\%}) &
  \textbf{37.5 (\textcolor{myyellow}{+7.5})} & \textbf{6948 (\textcolor{myblue}{-37\%})} &
  -- & -- \\
\multirow{-8}{*}{\textbf{DS-Qwen-7B}} & \cellcolor{gray!10}\textbf{\method} &
\cellcolor{gray!10}89.3 (\textcolor{myyellow}{-0.6}) & \cellcolor{gray!10}\textbf{385 (\textcolor{myblue}{-28\%})} &
\cellcolor{gray!10}\textbf{92.8 (\textcolor{myyellow}{+0.0})} & \cellcolor{gray!10}2237 (\textcolor{myblue}{-37\%}) &
\cellcolor{gray!10}\textbf{56.7 (\textcolor{myyellow}{+6.7})} & \cellcolor{gray!10}7978 (\textcolor{myblue}{-37\%}) &
\cellcolor{gray!10}36.7 (\textcolor{myyellow}{+6.7}) & \cellcolor{gray!10}9715 (\textcolor{myblue}{-12\%}) &
\cellcolor{gray!10}\textbf{92.5 (\textcolor{myyellow}{+5.0})} & \cellcolor{gray!10}3770 (\textcolor{myblue}{-41\%}) \\
\midrule

& COT & 94.9 & 1122 & 94.4 & 3539 & 60.0 & 10343 & 36.7 & 11002 & \textbf{92.5} & 5333 \\
& DEER & 93.3 (\textcolor{myyellow}{-1.6}) & 1040 (\textcolor{myblue}{-7\%}) &
        89.8 (\textcolor{myyellow}{-4.6}) & 2577 (\textcolor{myblue}{-27\%}) &
        \textbf{68.4 (\textcolor{myyellow}{+8.4})} & 8115 (\textcolor{myblue}{-22\%}) &
        36.7 (\textcolor{myyellow}{+0.0}) & 10125 (\textcolor{myblue}{-8\%}) &
        85.0 (\textcolor{myyellow}{-7.5}) & \textbf{4240 (\textcolor{myblue}{-20\%})} \\
& ThinkSwitcher & 94.3 (\textcolor{myyellow}{-0.6}) & 1042 (\textcolor{myblue}{-7\%}) &
        92.7 (\textcolor{myyellow}{-1.7}) & 3572 (\textcolor{myblue}{+1\%}) &
        60.4 (\textcolor{myyellow}{+0.4}) & \textbf{8044 (\textcolor{myblue}{-22\%})} &
        42.5 (\textcolor{myyellow}{+5.8}) & 10065 (\textcolor{myblue}{-9\%}) &
        -- & -- \\
\multirow{-4}{*}{\textbf{DS-Qwen-14B}} & \cellcolor{gray!10}\textbf{\method} &
\cellcolor{gray!10}\textbf{95.2 (\textcolor{myyellow}{+0.3})} & \cellcolor{gray!10}\textbf{621 (\textcolor{myblue}{-45\%})} &
\cellcolor{gray!10}\textbf{94.8 (\textcolor{myyellow}{+0.4})} & \cellcolor{gray!10}\textbf{2515 (\textcolor{myblue}{-29\%})} &
\cellcolor{gray!10}60.0 (\textcolor{myyellow}{+0.0}) & \cellcolor{gray!10}8393 (\textcolor{myblue}{-19\%}) &
\cellcolor{gray!10}\textbf{50.0 (\textcolor{myyellow}{+13.3})} & \cellcolor{gray!10}\textbf{8901 (\textcolor{myblue}{-19\%})} &
\cellcolor{gray!10}\textbf{92.5 (\textcolor{myyellow}{+0.0})} & \cellcolor{gray!10}4656 (\textcolor{myblue}{-13\%}) \\
\midrule

& COT & \textbf{90.2} & 1400 & 93.8 & 3412 & 56.7 & 10280 & \textbf{40} & 10893 & 90 & 5997 \\

& DEER & 89.8 (\textcolor{myyellow}{-0.4}) & 1473 (\textcolor{myblue}{+5\%}) &
91.4 (\textcolor{myyellow}{-2.4}) & 2995 (\textcolor{myblue}{-12\%}) & \textbf{66.7 (\textcolor{myyellow}{+10})} & 9755 (\textcolor{myblue}{-5\%}) & 36.7 (\textcolor{myyellow}{-3.3}) & 11820 (\textcolor{myblue}{+9\%}) & 90 (\textcolor{myyellow}{+0.0}) & 5408 (\textcolor{myblue}{-10\%}) \\

\multirow{-3}{*}{\textbf{Nemo-Llama-8b}} & \cellcolor{gray!10}\textbf{\method} &
\cellcolor{gray!10}89.7 (\textcolor{myyellow}{-0.5}) & \cellcolor{gray!10}\textbf{1035 (\textcolor{myblue}{-26\%})} &
\cellcolor{gray!10}\textbf{94 (\textcolor{myyellow}{+0.2})} & \cellcolor{gray!10}\textbf{2844 (\textcolor{myblue}{-17\%})} &
\cellcolor{gray!10}60 (\textcolor{myyellow}{+3.3}) & \cellcolor{gray!10}\textbf{9258 (\textcolor{myblue}{-10\%})} &
\cellcolor{gray!10}\textbf{40 (\textcolor{myyellow}{+0.0})} & \cellcolor{gray!10}\textbf{10400 (\textcolor{myblue}{-5\%})} &
\cellcolor{gray!10}\textbf{95 (\textcolor{myyellow}{+5})} & \cellcolor{gray!10}\textbf{4441 (\textcolor{myblue}{-26\%})} \\
\midrule

& COT & \textbf{97.0} & 1561 & \textbf{97.0} & 4025 & 66.7 & 11305 & \textbf{63.3} & 12554 & 90.0 & 7086 \\
& TALE & -- & -- &
  94.0 (\textcolor{myyellow}{-3.0}) & 3533 (\textcolor{myblue}{-12\%}) &
  61.1 (\textcolor{myyellow}{-5.6}) & 10888 (\textcolor{myblue}{-4\%}) &
  -- & -- &
  90.8 (\textcolor{myyellow}{+0.8}) & 6522 (\textcolor{myblue}{-8\%}) \\
& NoThinking & -- & -- &
  94.2 (\textcolor{myyellow}{-2.8}) & 4276 (+6\%) &
  62.2 (\textcolor{myyellow}{-4.5}) & 11688 (+3\%) &
  -- & -- &
  88.3 (\textcolor{myyellow}{-1.7}) & 7493 (+6\%) \\
& Dynasor & -- & -- &
  94.0 (\textcolor{myyellow}{-3.0}) & 4156 (+3\%) &
  64.4 (\textcolor{myyellow}{-2.3}) & 9733 (\textcolor{myblue}{-14\%}) &
  -- & -- &
  90.0 (\textcolor{myyellow}{+0.0}) & 7185 (+1\%) \\
& DEER & 96.3 (\textcolor{myyellow}{-0.7}) & 977 (\textcolor{myblue}{-37\%}) &
  94.6 (\textcolor{myyellow}{-2.4}) & 3316 (\textcolor{myblue}{-18\%}) &
  \textbf{70.0 (\textcolor{myyellow}{+3.3})} & 10097 (\textcolor{myblue}{-11\%}) &
  50.0 (\textcolor{myyellow}{-13.3}) & 11598 (\textcolor{myblue}{-8\%}) &
  \textbf{95.0 (\textcolor{myyellow}{+5.0})} & 5782 (\textcolor{myblue}{-18\%}) \\
& CGRS & -- & -- &
  94.2 (\textcolor{myyellow}{-2.8}) & \textbf{2810 (\textcolor{myblue}{-30\%})} &
  68.9 (\textcolor{myyellow}{-1.1}) & \textbf{8202 (\textcolor{myblue}{-27\%})} &
  -- & -- &
  93.3 (\textcolor{myyellow}{+3.3}) & \textbf{4771 (\textcolor{myblue}{-33\%})} \\
\multirow{-7}{*}{\textbf{QwQ-32B}} & \cellcolor{gray!10}\textbf{\method} &
\cellcolor{gray!10}96.6 (\textcolor{myyellow}{-0.4}) & \cellcolor{gray!10}\textbf{969 (\textcolor{myblue}{-38\%})} &
\cellcolor{gray!10}\textbf{97.0 (\textcolor{myyellow}{+0.0})} & \cellcolor{gray!10}3256 (\textcolor{myblue}{-19\%}) &
\cellcolor{gray!10}\textbf{70.0 (\textcolor{myyellow}{+3.3})} & \cellcolor{gray!10}9181 (\textcolor{myblue}{-19\%}) &
\cellcolor{gray!10}53.3 (\textcolor{myyellow}{-10.0}) & \cellcolor{gray!10}\textbf{11416 (\textcolor{myblue}{-9\%})} &
\cellcolor{gray!10}\textbf{95.0 (\textcolor{myyellow}{+5.0})} & \cellcolor{gray!10}5777 (\textcolor{myblue}{-18\%}) \\
\bottomrule[1.5pt]
\end{tabular}
}
\caption{{Comparisons with additional baselines.} 
Metrics include \textbf{Acc} ($\uparrow$) and \textbf{Tokens} ($\downarrow$). Changes relative to the COT are highlighted in \textcolor{myyellow}{orange} for Acc and \textcolor{myblue}{blue} for Tokens. Best results within each group are bolded.}
\label{tab:more_baselines_results}
\end{table*}

\subsection{Process-Level Behavior Analysis on MATH500}
\label{app:exp_process_behavior}

\begin{figure}[t]
    \centering
    \includegraphics[width=\linewidth]{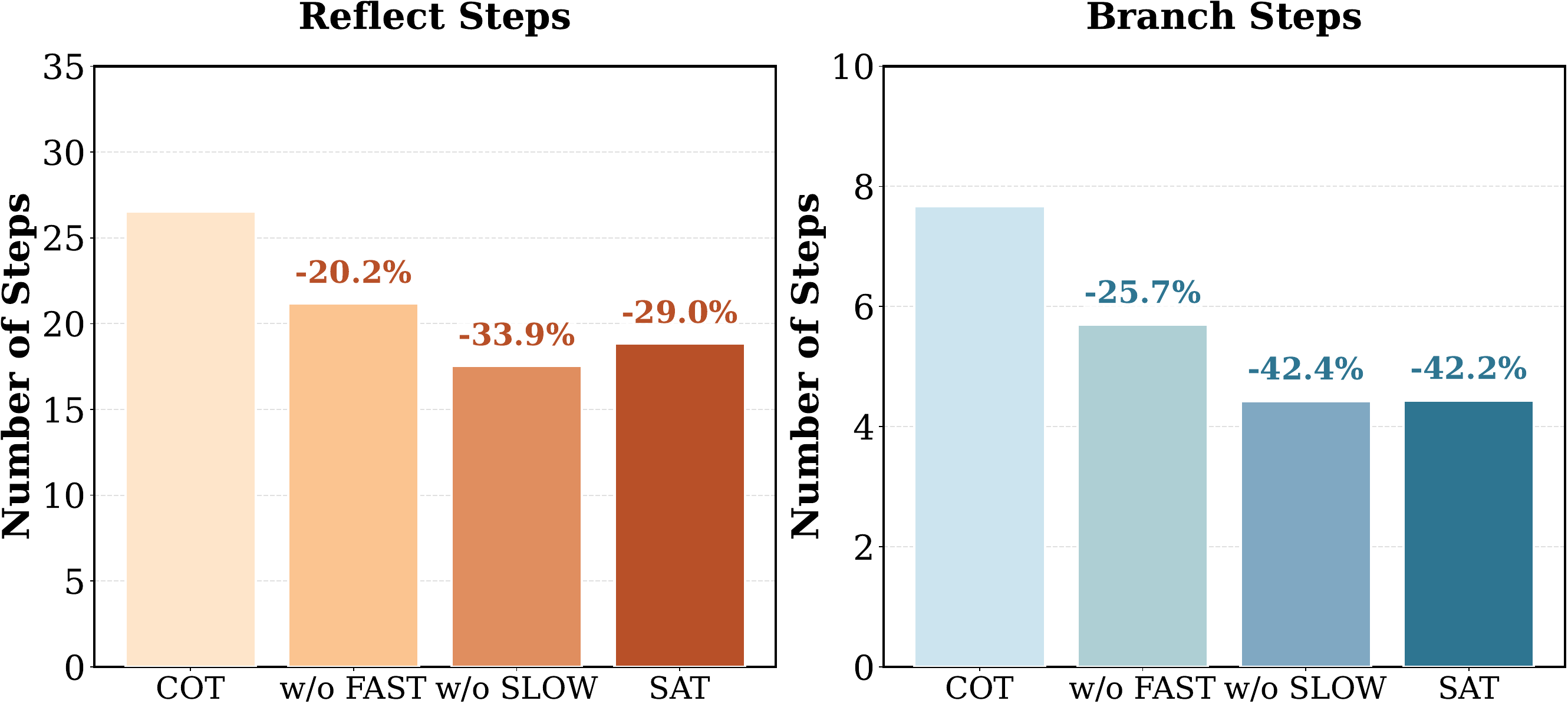}
    \caption{{Process behavior analysis on MATH500.} Average \texttt{reflect\_steps} and \texttt{branch\_steps} (lower is better) for \textit{COT}, \textit{w/o FAST}, \textit{w/o SLOW}, and \method, computed from the thought traces using the forward window scheme ($W{=}1$).}
    \label{fig:math500_reflect_steps}
    \vspace{-0.3cm}
\end{figure}

\paragraph{Definition of reflection/branching cues.}
We analyze the \emph{thought part} (text before \texttt{</think>}) and split it into step units by newline delimiters.
A \emph{reflection cue} is a lexical indicator that the model is verifying, reconsidering, or correcting its reasoning.
Typical examples include phrases such as \textit{``check''}, \textit{``verify''}, \textit{``wait''}, or \textit{``actually''}.
Similarly, a \emph{branching cue} indicates explicit exploration of alternatives, e.g., \textit{``case 1/2''}, \textit{``another approach''}, or \textit{``on the other hand''}.
We use a fixed cue list shared across all methods for fairness.

\paragraph{Computation of \texttt{reflect\_steps} and \texttt{branch\_steps}.}
We first mark a step as a cue-hit if it contains any cue term.
Under the forward window scheme (default $W{=}1$), if step $i$ is a cue-hit, then steps $i$ through $i{+}W$ are marked as belonging to a reflection (or branching) segment.
\texttt{reflect\_steps} and \texttt{branch\_steps} are defined as the total number of marked steps for reflection and branching, respectively.

\paragraph{Results.}
Figure~\ref{fig:math500_reflect_steps} shows that both \method and the ablated variants reduce reflection and branching behaviors compared to COT.
COT exhibits $26.55$ \texttt{reflect\_steps} and $7.67$ \texttt{branch\_steps} on average.
\textit{w/o FAST} reduces \texttt{reflect\_steps} to $21.20$ (\(-20.2\%\)) and \texttt{branch\_steps} to $5.70$ (\(-25.7\%\)).
\textit{w/o SLOW} yields the strongest suppression (reflect: $17.54$, \(-33.9\%\); branch: $4.42$, \(-42.4\%\)).
The full \method achieves comparable reductions (reflect: $18.86$, \(-29.0\%\); branch: $4.43$, \(-42.2\%\)), indicating that stepwise mode navigation can effectively curb redundant reflection and excessive branching.

\subsection{Sensitivity of attribution metrics to window size}
\label{app:why_short_window}

\begin{figure}[t]
    \centering
    \includegraphics[width=\linewidth]{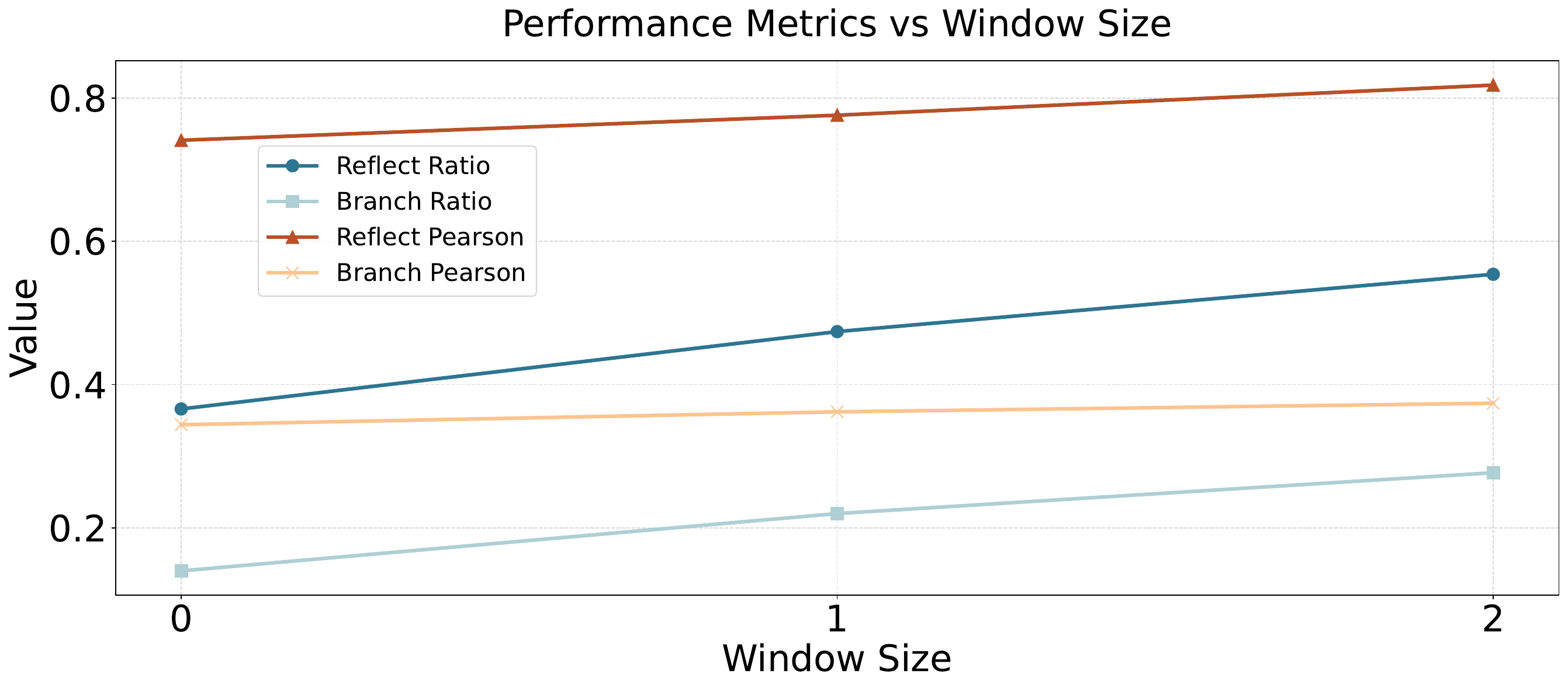}
    \caption{{Window-size sensitivity for token-savings attribution.} We report (i) contribution ratios of reflection-/branch-related savings and (ii) Pearson correlations between per-sample total savings and each component, under window sizes \(W\in\{0,1,2\}\).}
    \label{fig:appendix_why_short}
    \vspace{-0.3cm}
\end{figure}

\begin{figure}[t]
    \centering
    \includegraphics[width=0.9\linewidth]{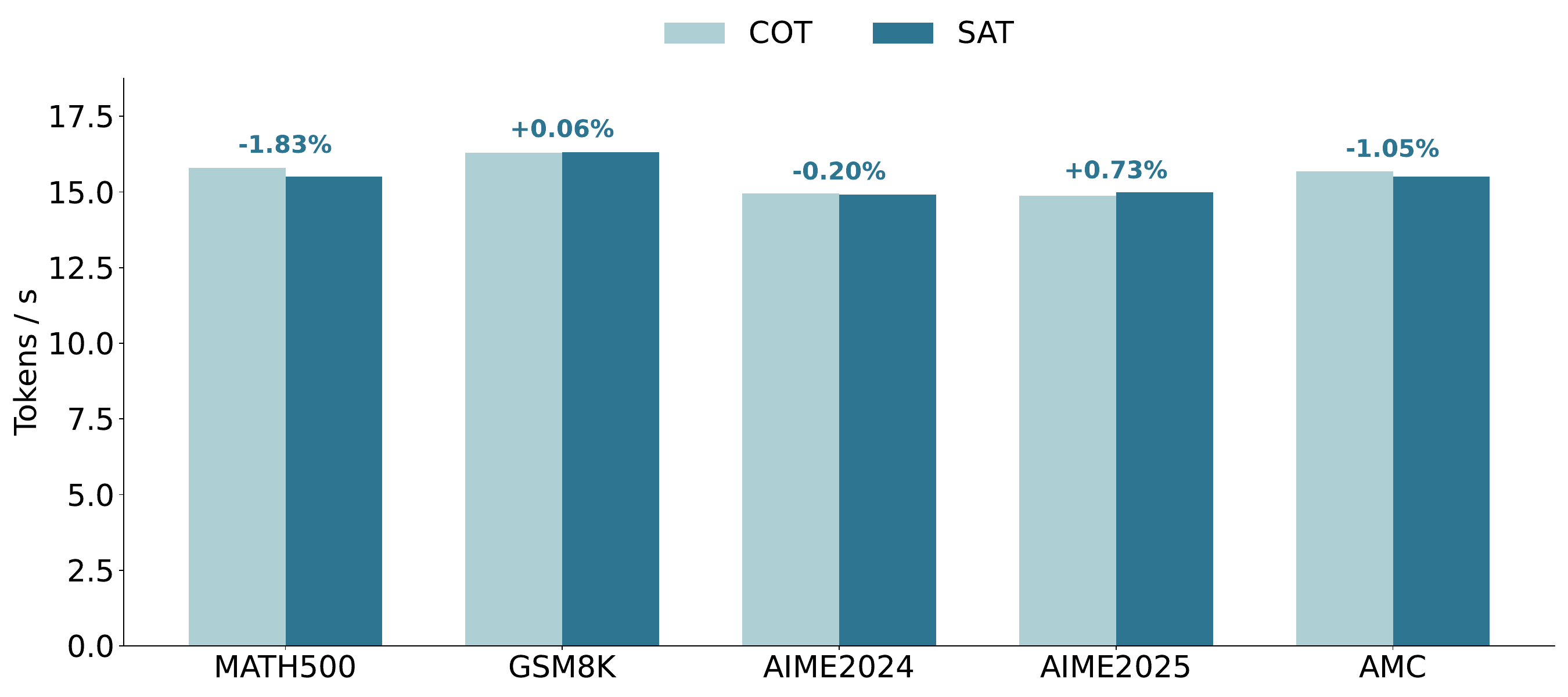} 
    \caption{{Generation Throughput Comparison.} The average number of tokens generated per second by COT and \method using Qwen3-32B. The consistent throughput confirms that \method introduces negligible per-step latency overhead.}
    \label{fig:throughput_comparison}
\end{figure}

\paragraph{How the attribution numbers are computed.}
We attribute \method's token savings to two process behaviors, \emph{reflection} and \emph{branching}, computed on the thought trace (text before \texttt{</think>}) split into step units by newline.
We first detect \emph{cue steps} using a fixed lexical list shared across all methods (e.g., reflection cues such as \texttt{``check''}, \texttt{``verify''}, \texttt{``wait''}, \texttt{``actually''}).
To better capture short cue-triggered spans, we apply a forward window expansion: if step \(i\) contains a reflection cue, then steps \(i,\ldots,i{+}W\) are marked as reflection-related, and the reflection token count is the sum of tokens over marked steps.
For a pair of methods (COT vs.\ \method), we define the \emph{reflection contribution ratio} as the fraction of total token savings (computed in a positive-only manner to avoid cancellation) that can be explained by the reduction in reflection-marked tokens; we further compute the Pearson correlation between per-sample total savings and per-sample reflection-marked savings.
Branch-related ratios and correlations are computed analogously, using branching cues (e.g., \texttt{``case''}, \texttt{``alternative''}, \texttt{``another way''}).

\paragraph{Why we choose \(W{=}2\), and robustness.}
We use \(W{=}2\) in the main text because it provides a slightly more inclusive segmentation of cue-triggered spans, yielding a clearer attribution signal while remaining conservative (only a short local expansion).
As shown in Figure~\ref{fig:appendix_why_short}, the qualitative conclusions are stable for smaller windows:
even with \(W{=}1\), reflection remains the dominant contributor (reflect\_ratio \(0.474\) vs.\ branch\_ratio \(0.220\)) and is strongly aligned with total savings (reflect Pearson \(0.776\) vs.\ branch Pearson \(0.362\)).
Increasing to \(W{=}2\) strengthens the same trend (reflect\_ratio \(0.554\), reflect Pearson \(0.818\)), supporting the interpretation that reduced reflection is the primary driver of \method's token efficiency, with branching reduction as a secondary factor.

\subsection{Throughput Analysis}
\label{sec:appendix_throughput}

To rigorously assess the computational overhead introduced by the external Pilot and FSM logic, we compare the generation throughput (measured in tokens per second) of \method against vanilla COT. 
The evaluation is conducted using \textbf{Qwen3-32B} across five mathematical benchmarks: MATH500, GSM8K, AIME 2024, AIME 2025, and AMC.

As illustrated in Figure~\ref{fig:throughput_comparison}, the generation speed of \method is virtually indistinguishable from that of vanilla COT across all datasets (e.g., \textbf{15.51} vs.\ \textbf{15.68} tokens/s on AMC; \textbf{16.30} vs.\ \textbf{16.30} tokens/s on GSM8K). 
This parity indicates that the lightweight 30M Pilot operates with negligible latency, likely fully masked by the memory-bound decoding process of the 32B backbone. 
Consequently, the end-to-end speedup reported in the main text is purely derived from the reduction in total generated tokens, validating that \method incurs no throughput penalty during inference.

\section{Potential Risks}
\label{app:risks}

While our proposed framework, Stepwise Adaptive Thinking (SAT), significantly improves the inference efficiency of Large Reasoning Models (LRMs), we acknowledge the broader risks inherent to AI systems that deploy such techniques:

\paragraph{Reliability and Hallucination.} 
Although our method aims to preserve accuracy by retaining verification steps for complex problems, LRMs are fundamentally probabilistic and prone to hallucinations. The dynamic pruning of reasoning steps (e.g., in \textit{Fast} or \textit{Skip} modes) introduces a trade-off where subtle, necessary self-corrections might occasionally be bypassed, potentially leading to confident but incorrect outputs in out-of-distribution scenarios.

\paragraph{Misuse of Efficient Inference.} 
By reducing the computational barrier and token cost of complex reasoning (achieving up to 40\% reduction), our work unintentionally lowers the cost for malicious actors to deploy high-capability models at scale. This could facilitate the automated generation of sophisticated disinformation, social engineering attacks, or malicious code at a lower economic cost.

We emphasize that human oversight remains critical, especially in high-stakes domains, to ensure that the pursuit of efficiency does not compromise the safety and ethical integrity of the reasoning process.

\section{Use of AI Assistants}
\label{app:ai_use}

AI assistants were used in a limited and supportive role during the writing process, primarily for language polishing and text refinement.
\end{document}